\pdfoutput=1

\documentclass[11pt]{article}

\usepackage[final]{latex/acl}

\usepackage{times}
\usepackage{latexsym}

\usepackage[T1]{fontenc}

\usepackage[utf8]{inputenc}
\usepackage{times}
\usepackage{booktabs}
\usepackage{authblk}
\usepackage{latexsym}
\usepackage{multirow}
\usepackage{amssymb}
\usepackage{float}
\usepackage{graphicx}
\usepackage{amsmath}
\usepackage{algorithm}
\usepackage[noend]{algpseudocode}
\usepackage{tablefootnote}
\usepackage{adjustbox}
\usepackage[most]{tcolorbox}
\usepackage{colortbl}
\usepackage{seqsplit}
\newtcolorbox{mybox}[2][]{
    colback=white,
    colframe=green!45,
    fonttitle=\bfseries,
    coltitle=black,
    sharp corners,
    title=#2,
    #1
}
\usepackage{url}
\usepackage{wrapfig}
\usepackage{graphics}
\usepackage{bbm}
\usepackage{amsmath,amsfonts,bm}
\usepackage{makecell}
\usepackage{array}
\usepackage[para]{threeparttable}
\usepackage{booktabs}
\usepackage{tabularx}
\usepackage{xspace}
\usepackage{tablefootnote}
\usepackage{adjustbox}
\usepackage{subcaption}
\usepackage{microtype}

\usepackage{inconsolata}
\usepackage[most]{tcolorbox}
\usepackage{colortbl}
\title{\includegraphics[width=0.550cm, height=0.637cm]{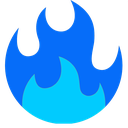}~\textsc{Prometheus-Vision}:\\ Vision-Language Model as a Judge for Fine-Grained Evaluation}

\makeatletter
\renewcommand\AB@affilsepx{\hspace{0.5cm} \protect\Affilfont}
\makeatother

\author{\textbf{Seongyun Lee}$^{1}$\thanks{~denotes equal contribution. Work was done when Seongyun was an intern at KAIST AI and Seungone was an intern at NAVER AI Lab.} \quad \textbf{Seungone Kim}$^{1,2}$$\textbf{}^{*}$ \quad \textbf{Sue Hyun Park}$^{1}$ \quad \textbf{Geewook Kim}$^{1,3}$ \quad \textbf{Minjoon Seo}$^{1}$ \\ 
\\
KAIST AI$^{1}$ \quad NAVER AI Lab$^{2}$ \quad NAVER Cloud AI$^{3}$\\
\texttt{\{seongyun, seungone, suehyunpark, geewook, minjoon\}@kaist.ac.kr}} 

\begin{document}
\maketitle
\begin{abstract}
Assessing long-form responses generated by Vision-Language Models (VLMs) is challenging. It not only requires checking whether the VLM follows the given instruction but also verifying whether the text output is properly grounded on the given image. Inspired by the recent approach of evaluating LMs with LMs, in this work, we propose to evaluate VLMs with VLMs. For this purpose, we present a new feedback dataset called the \textsc{Perception Collection}, encompassing 15K customized score rubrics that users might care about during assessment. Using the \textsc{Perception Collection}, we train \textsc{Prometheus-Vision}, the first open-source VLM evaluator model that can understand the user-defined score criteria during evaluation. \textsc{Prometheus-Vision} shows the highest Pearson correlation with human evaluators and GPT-4V among open-source models, showing its effectiveness for transparent and accessible evaluation of VLMs. We open-source our code, dataset, and model at \url{https://github.com/kaistAI/prometheus-vision}.
\end{abstract}

\section{Introduction}

\begin{figure*}[!t]
\centering
  \noindent\includegraphics[width=\linewidth]{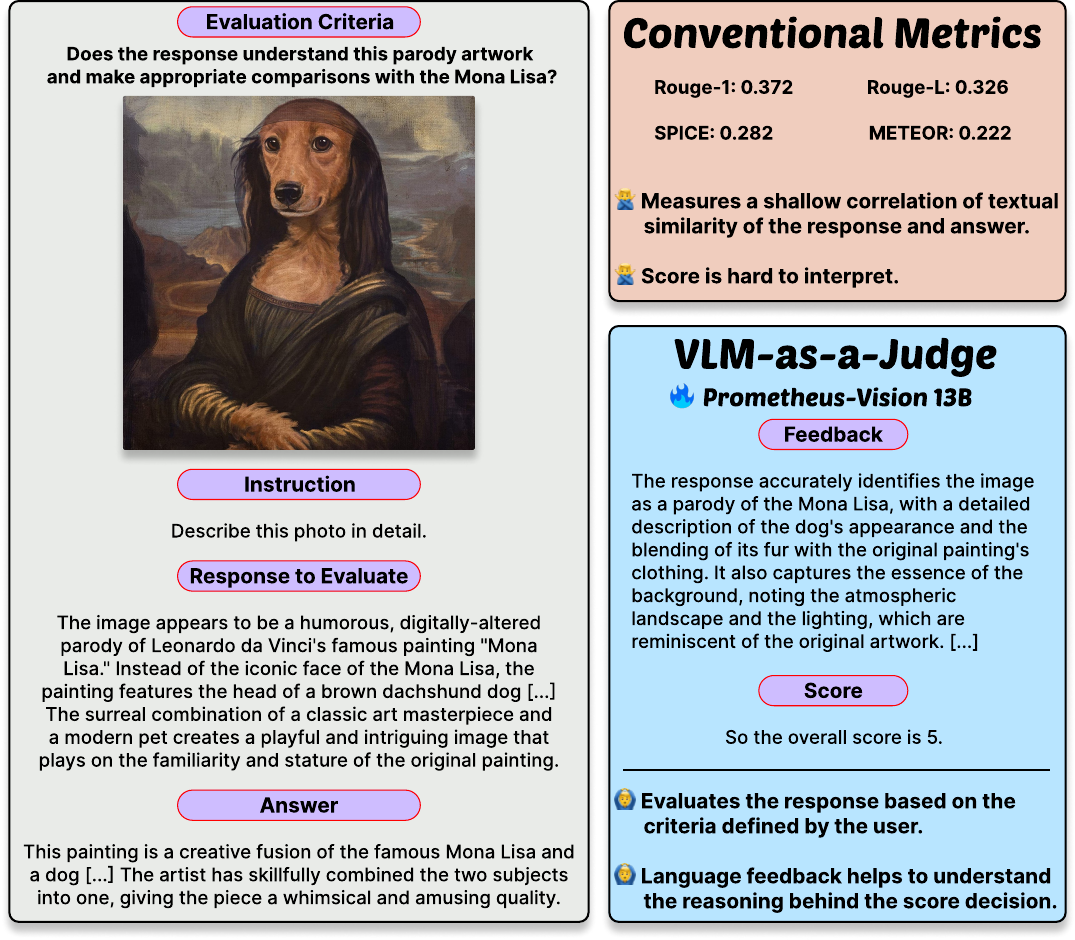} 
  \caption{Conventional metrics measure the similarity between the response and ground-truth answer, which is not expressive enough. Moreover, it could not pinpoint what is missing within the response with respect to the evaluation criteria. In contrast, the \textbf{VLM-as-a-Judge} pipeline provides not only the flexibility to adhere to arbitrary evaluation criteria but also provides detailed language feedback that specifically pinpoints the deficiencies.}
  \label{fig:conventional_vlmjudge}
\end{figure*}

While recently developed Vision-Language Models (VLMs) are capable of generating long-form text from a combination of an image and instruction, assessing the quality of the output remains a significant challenge~\citep{liu2023improved, dai2023instructblip, gao2023llama, ye2023mplug, zhu2023minigpt, gpt4-v}. Traditional metrics, which rely on text-based exact matches or edit distances, fall short in adhering to the granular evaluation criterion of interest and capturing the rich context within the outputs~\citep{agrawal2023reassessing,manas2023improving,bai2023touchstone}. For instance, as shown in Figure~\ref{fig:conventional_vlmjudge}, conventional metrics fail to explain what is missing within the response compared to the answer.


Consequently, the role of high-quality human evaluations remains pivotal for a comprehensive assessment. However, human evaluators are prone to biases, and scaling up is expensive in terms of time and cost~\citep{ye2023flask,kim2023cotever}.

To address the need for flexible and automatic text evaluation, the `LM-as-a-Judge' paradigm proposes using language models (LMs) as evaluators, where initial findings suggest its potential to emulate human judgement~\citep{liu2023improved,zheng2023judging, alpaca_eval, ye2023flask, kim2023prometheus, zhu2023judgelm, bai2023touchstone}.
However, LMs cannot perceive visual contexts, which necessitates an additional model that could convert the image to text. As a result, such a multistage pipeline could potentially suffer from error propagation and also require multiple inference calls.
This situation calls for the direct utilization of VLMs, referred to as \textbf{VLM-as-a-Judge}.

On the other hand, despite GPT-4V's~\citep{gpt4-v} potential as an evaluator, its closed-source nature limits transparent evaluation~\citep{kim2023prometheus}. Moreover, our initial tests indicate that open-source VLMs are not effective as evaluators sensitive to granular aspects, demonstrating a low score correlation with both human evaluators and GPT-4V. To address these challenges, we propose \textsc{Prometheus-Vision}, a 13B VLM evaluator that excels at assessing based on fine-grained criteria. As shown in Figure~\ref{fig:conventional_vlmjudge}, \textsc{Prometheus-Vision} could evaluate based on the given criteria, pinpointing the differences between the parody artwork and the original masterpiece.

To develop \textsc{Prometheus-Vision}, we construct the \textsc{Perception Collection}, the first multi-modal feedback dataset that includes 15K fine-grained score rubrics, thus going beyond traditional coarse-grained criteria such as helpfulness, relevance, accuracy, and comprehensiveness. Using the \textsc{Perception Collection}, we fine-tune LLaVA-1.5 to create \textsc{Prometheus-Vision}. 

We achieve our objective by fine-tuning a VLM based on fine-grained criteria with reference materials (\textit{e.g.}, reference answer, score rubric) appended. The resulting evaluator VLM, \textsc{Prometheus-Vision}, shows a high correlation with both human evaluators and GPT-4V, serving as a testament to its efficiency. The proposed system exhibits a Pearson correlation of 0.786 with human evaluators and 0.639 with GPT-4V on the LLaVA-Bench. These results highlight its potential to serve as an inexpensive yet effective open-source alternative to GPT-4V evaluation. Moreover, when assessing 3,560 instances across 8 benchmarks, \textsc{Prometheus-Vision} shows the highest correlation with GPT-4V on all the 8 benchmarks among the open-source models and even outperforms GPT-4 (LM judge) on 5 benchmarks.


Our contributions are summarized as follows:
\begin{itemize}
    \item We introduce \textsc{Perception Collection}, the first multi-modal feedback dataset that could be used to train an evaluator VLM. In contrast to existing multi-modal feedback, critique, and preference datasets that use coarse-grained criteria, the \textsc{Perception Collection} includes 15K fine-grained criteria that determine the crucial aspect for each instance.
    \item We introduce \textsc{Prometheus-Vision}, the first open-source VLM specialized for evaluation purposes. \textsc{Prometheus-Vision} shows a high correlation with both GPT-4V and human evaluators, indicating its potential to be used as a cheap alternative for GPT-4V evaluation.
\end{itemize}



\section{Related works}

\subsection{Evaluating Vision Language Models} 


In prior works, Vision-Language Models (VLMs) are typically evaluated using specific metrics tailored to each task. For image captioning, performance is measured with metrics like BLEU~\citep{papineni-etal-2002-bleu}, METEOR~\citep{banerjee-lavie-2005-meteor}, ROUGE~\citep{lin2004rouge}, and CIDEr~\citep{Vedantam_2015_CVPR}, focusing on how well the generated text aligns with reference captions. Similarly, Visual Question Answering (VQA) is evaluated using accuracy metrics based on the exact match between the model's answers and human-annotated answers~\citep{agrawal2023reassessing,manas2023improving}. 

However, traditional metrics often fall short of capturing the nuanced details of the response generated by VLMs in complex or subjective situations. A more comprehensive approach has been human evaluation, accounting for contextual and creative aspects not captured by automated metrics. Nonetheless, cost and consistency constraints associated with human evaluations render it a less feasible method for scaling to a lot of instances. 

\subsection{Language Model as a Judge for Fine-grained Evaluation}
The difficulty in evaluating long-form responses often arises from the ambiguity in defining what constitutes a good output. For instance, discerning whether a given response is helpful or harmless is often subjective. Recent works have proposed the concept of `Fine-grained Evaluation', utilizing LM-as-a-judge for assessing granular aspects. \citet{ye2023flask} defines 12 core skill sets that are crucial for evaluating LMs. \citet{kim2023prometheus} further extends this concept and employs thousands of fine-grained criteria to assess LMs on user-defined criteria. \citet{wu2023finegrained} and \citet{jang2023personalized} utilize fine-grained criteria to align LMs. Lastly, \citet{kim2023evallm} proposes an interactive framework in which users could test LMs on fine-grained criteria.

To the best of our knowledge, we are first to expand the notion of `Fine-grained Evaluation' for assessing VLMs. Specifically, recent work has proposed to evaluate VLMs using LMs or VLMs~\citep{bai2023touchstone,ge2023mllm}, yet are still confined to high-level coarse-grained criteria such as helpfulness, relevance, accuracy, and comprehensiveness. We construct the \textsc{Perception Collection} which encompasses 15K of fine-grained criteria and use it to train \textsc{Prometheus-Vision}.

\begin{figure*}[!t]
\centering
  \noindent\includegraphics[width=\linewidth]{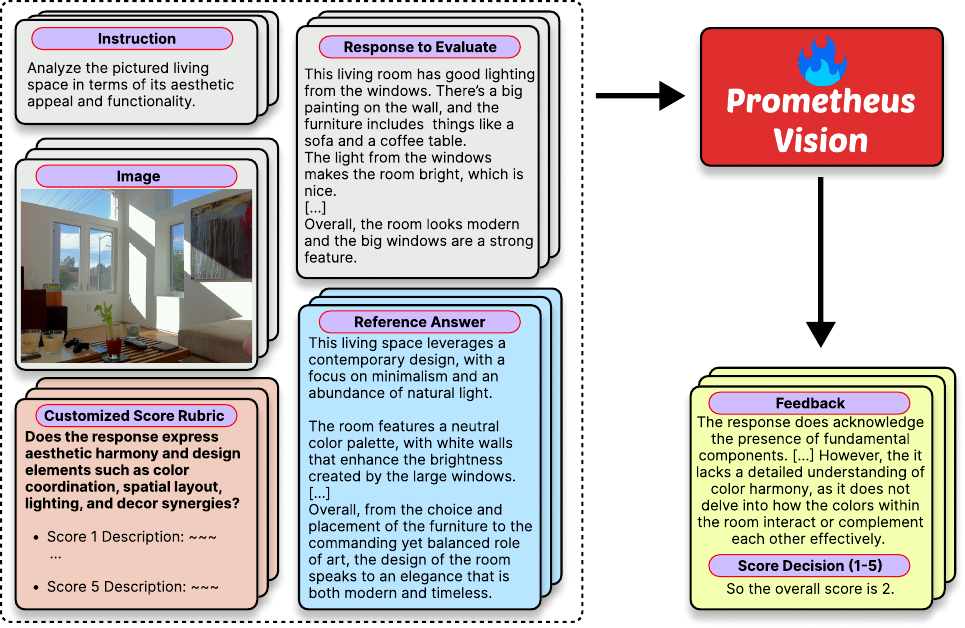} 
  \caption{Previous automatic metrics could not capture whether a VLM's response is aware of \textit{aesthetic harmony}. With \textsc{Prometheus-Vision}, users could define customized score rubrics that they care about instead of assessing based on coarse-grained criteria such as helpfulness, relevance, accuracy, and comprehensiveness. Each component within the \textsc{Perception Collection} consists of 5 input components: an instruction, a real-world image, a response to evaluate, a customized score rubric, and a reference answer. Based on this, \textsc{Prometheus-Vision} is trained to generate a language feedback and a score decision.}
  \label{fig:component_pipeline}
\end{figure*}

\section{The \textsc{Perception Collection}}


In contrast to the language domain, to the best of our knowledge, there do not exist any available feedback, critique, or preference datasets applicable to train an evaluator VLM that could assess in a \textit{fine-grained} manner. For this purpose, we first construct a comprehensive multi-modal feedback dataset called the \textsc{Perception Collection}.

As shown in Figure~\ref{fig:component_pipeline}, each instance in the \textsc{Perception Collection} consists of five input components (image, instruction, response to evaluate, customized score rubric, reference answer) and two output components (language feedback and score decision). The number of each component in the \textsc{Perception Collection} is shown in Table~\ref{table:perception_collection_stats}.

Specifically, the five input components are:
\begin{itemize}
    \item \textbf{Image}: A real-world image that the user would provide to the VLM.
    \item \textbf{Instruction}: A text instruction that the user would prompt the VLM. It is also related to the provided image.
    \item \textbf{Response to Evaluate}: A text response that the VLM would generate based on the image and instruction. The evaluator VLM has to assess this response.
    \item \textbf{Customized Score Rubric}: A detailed scoring criteria that the VLM should refer to for assessment. We use fine-grained criteria in contrast to coarse-grained ones such as helpfulness, relevance, accuracy, and comprehensiveness. The rubric consists of (1) a description of the criteria and (2) a description of each scoring decision on a scale of 1 to 5.
    \item \textbf{Reference Answer}: A reference answer that would achieve a score of 5. While this component could be hand-crafted by human annotators, in our experiments, we utilize GPT-4V.
\end{itemize}

Moreover, the two output components are:
\begin{itemize}
    \item \textbf{Feedback}: A rationale pinpointing what is good and bad about the response under assessment. Instead of directly providing a scoring decision, this component makes the judgement process more interpretable.
    \item \textbf{Score}: An integer value on a scale of 1 to 5 that represents the quality of the response given the criteria mentioned in the score rubric.
\end{itemize}

\subsection{\textsc{Perception Collection} Construction}\label{section:3.1}

\begin{table}
\centering
\fontsize{8}{10}\selectfont
\begin{tabular}{lcc}
    \toprule
    Components & \# Components & \# Components per Image\\
    \midrule
    Images&5,000& 1\\
    Score Rubrics&15,000&3\\
    Instructions&30,000&6\\
    Reference Answers&30,000&6\\ 
    Responses&150,000&30\\
    Feedback \& Score&150,000&30\\
    \bottomrule    
\end{tabular}
\caption{The number of each component included in the \textsc{Perception Collection}. Note that the feedback and score are evenly distributed, leading to 30K instances per score between 1 and 5.}
\label{table:perception_collection_stats}
\end{table}  

We construct a multi-modal feedback dataset called the \textsc{Perception Collection}. We mainly follow the construction process of \citet{kim2023prometheus}. While creating the \textsc{Perception Collection}, we utilize 5K real-world images sampled from MS COCO 2017 Challenge~\citep{lin2014microsoft} and the MMMU benchmark~\citep{yue2023mmmu}. 

Concretely, the augmentation process consists of 4 stages: (1) hand-crafting 50 seed score rubrics, (2) brainstorming 15K fine-grained score rubrics, (3) augmenting 30K instructions and reference answers closely tied with the rubric, and (4) augmenting 150K responses and language feedback. We include a detailed analysis of the \textsc{Perception-Collection} in terms of diversity and quality in Appendix~\ref{section:A} and all the prompts used for augmentation in Appendix~\ref{section:E}.

\paragraph{Step 1: Hand-Crafting Score Rubrics} We first start by writing 50 examples of fine-grained score rubrics that go beyond the coarse-grained counterparts. For 50 images, we write an instruction and the corresponding rubric that pinpoints which aspect to consider during the assessment.

\paragraph{Step 2: Brainstorming Score Rubrics} Using GPT-4V, we expand the number of our score rubrics from 50 to 15K. Using an arbitrary image among the 5K pool and the 50 examples as demonstrations, we prompt GPT-4V to generate 3 variants for each image. To ensure quality, we go through an additional stage of prompting GPT-4V to inspect whether the generated score rubric \textit{aligns} with the image. If it does not, we iteratively prompt it again until we acquire 3 candidates per image.

\paragraph{Step 3: Augmenting Instructions and Reference Answers related to the Score Rubric} Afterwards, we use the 15K score rubrics and prompt GPT-4V to generate 2 novel instructions for each score rubric, leading to a total number of 30K. This process ensures that the instruction is closely tied to the score rubric since the instruction was conditioned on the score rubric.

\paragraph{Step 4: Augmenting Training Instances} Lastly, we augment the remaining components which are the response to evaluate, feedback, and scoring decision. We use the score rubric and instruction generated from the previous stages and prompt GPT-4V to write a response that would get a score of $i$ (1 $\leq$ $i$ $\leq$ 5). Importantly, we ensured that there is no length bias (\textit{i.e.}, giving a higher score for longer responses) and included an analysis at Section~\ref{section:6.2}. This leads to a total number of 150K responses and 150K feedback where each score within between 1 and 5 has an even number of 30K instances. 

We include our analysis of the \textsc{Perception Collection} in terms of its quality, diversity, and whether there is a length bias among score decisions at Appendix~\ref{section:A}.



\subsection{Fine-tuning a VLM as an Evaluator}\label{section:3.2}

Using the \textsc{Perception Collection}, we use LLaVA-1.5 (7B \& 13B)~\citep{liu2023improved} as our backbone model and train \textsc{Prometheus-Vision} (7B \& 13B). Training on the \textsc{Prometheus Collection} is analogous to Chain-of-Thought fine-tuning which requires generating a rationale (which is the feedback in our case) and then the score in a sequential manner~\citep{ho2022large,kim2023cot}. We include a fixed phrase `\texttt{So the overall score is}' in between the feedback and the score which we found to prevent degeneration during inference. The detailed hyper-parameters used during training are included in Appendix~\ref{section:C.1}.

\section{Experimental Settings}
\subsection{Protocol for Evaluating Evaluator VLMs}\label{section:4.1}
In this section, we explain our experimental setting used to assess the fine-grained judgement capabilities of evaluator VLMs. As it is a non-trivial problem to directly measure `\textit{How well a VLM is evaluating}', we indirectly compare with two different standards: (1) how closely \textsc{Prometheus-Vision} could simulate human evaluators (Section~\ref{section:5.1}) and (2) how closely \textsc{Prometheus-Vision} could simulate the best VLM, which is GPT-4V, for nuanced assessment purposes (Section~\ref{section:5.2}). 

\subsection{Evaluator VLM \& LM Baselines}
We employ 9 VLMs as our evaluator VLM baselines, namely \textsc{LLaVA-1.5} (7B \& 13B)~\citep{liu2023improved}; \textsc{LLaVA-RLHF} (7B \& 13B)~\citep{sun2023aligning}; \textsc{ShareGPT4V} (7B)~\citep{chen2023sharegpt4v}; \textsc{Fuyu} (8B)~\citep{fuyu-8b}; and \textsc{GPT-4V}~\citep{gpt4-v} along with \textsc{Prometheus-Vision} (7B \& 13B).

In addition, we also compare with using LMs as a judge for evaluating VLMs as in previous work~\citep{bai2023touchstone}. We add 4 LMs as our evaluator LM baselines, namely \textsc{Prometheus} (7B \& 13B)~\citep{kim2023prometheus}; \textsc{GPT-3.5-Turbo}~\citep{chatgpt}; and \textsc{GPT-4}~\citep{openai2023gpt4}. Since LMs could not receive images as input, we prompt LLaVA-1.5 to generate a caption for the given image and provide the caption as additional input for LM evaluators. In contrast, for VLM evaluator baselines, we directly provide the image as input. The hyper-parameters used to inference evaluator LMs and evaluator VLMs are included in Appendix~\ref{section:C.1}.

\subsection{Response VLMs}
During our experiments, we utilize 3 different VLMs to sample the outputs that our VLM evaluators would assess. We denote these 3 VLMs as `Response VLMs'. We utilize \textsc{Fuyu} (8B), \textsc{LLaVA-1.5} (13B), and \textsc{GPT-4V} as our response VLM. The hyper-parameters used to inference response VLMs are included in Appendix~\ref{section:C.1}.

\subsection{Benchmarks}
Our evaluation benchmarks are mainly divided into 3 categories:
\begin{itemize}
    \item \textbf{Visual Instruction Following Benchmarks}: Tasks that require to write a long-form text output given an image and a text instruction. We use LLaVA-Bench~\citep{liu2023improved}, VisIT-Bench~\citep{bitton2023visit}, and a held-out test set of the \textsc{Perception Collection} called the \textsc{Perception Bench}.
    \item \textbf{Visual Question Answering Benchmarks}: Tasks that require to write a text output given an image and a text question. Compared to instruction following benchmarks, one notable difference is that we use the short-form answers originated from each dataset as reference answers in the input. We use the test set of the OKVQA dataset~\citep{marino2019ok}, VQAv2 dataset~\citep{goyal2017making}, and TextVQA dataset~\citep{singh2019towards}.
    \item \textbf{Captioning Benchmarks}: Tasks that require to write a text caption of the given image. Similar to the visual question answering benchmarks, the ground truth answers tend to be short compared to the reference responses in the instruction following benchmarks. We use the test set of the COCO-Captions dataset~\citep{chen2015microsoft} and NoCaps dataset~\citep{agrawal2019nocaps}.
\end{itemize}

\begin{table}
\centering
\fontsize{8}{10}\selectfont
\begin{tabular}{lcc}
    \toprule
    Benchmarks & \# Instances & \# Score Rubrics\\
    \midrule
    \textsc{LLaVA-Bench}&15& 15 (Hand-crafted)\\
    \textsc{VisIT-Bench}&15&15 (Hand-crafted)\\
    \textsc{Perception-Bench}&15&15 (Hand-crafted)\\
    \midrule
    \textsc{Total} & 45 & 45\\
    \bottomrule    
\end{tabular}
\caption{The number of the instances and score rubrics included in our evaluation setting in Section~\ref{section:5.1}. We randomly sample 15 instances from each benchmark and hand-craft a instance-wise fine-grained score rubric. Each instance originally has an image and an instruction.}
\label{table:benchmarks_stats_human}
\end{table}  

\begin{table}
\centering
\fontsize{7.5}{9}\selectfont
\begin{tabular}{lcc}
    \toprule
    Benchmarks & \# Instances & \# Score Rubrics\\
    \midrule
    \textsc{LLaVA-Bench}&60& 60 (Machine-generated)\\
    \textsc{VisIT-Bench}&500&500 (Machine-generated)\\
    \textsc{Perception-Bench}&500&500 (Machine-generated)\\
    \textsc{OKVQA}&500&5 (Machine-generated)\\ 
    \textsc{VQAv2}&500&5 (Machine-generated)\\
    \textsc{TextVQA}&500&5 (Machine-generated)\\
    \textsc{COCO-Captions}&500&5 (Machine-generated)\\
    \textsc{No-Caps}&500&5 (Machine-generated)\\
    \midrule
    \textsc{Total} & 3560 & 1085\\
    \bottomrule    
\end{tabular}
\caption{The number of the instances and score rubrics included in our evaluation setting in Section~\ref{section:5.2}. Except for LLaVA-Bench, we randomly sample 500 instances from each benchmark. Each instance originally has an image and an instruction. We additionally add a fine-grained score rubric and reference answer by prompting GPT-4V as explained in Section~\ref{section:3.1}.}
\label{table:benchmarks_stats_gpt4v}
\end{table}  

The number of instances and score rubrics for each benchmark is shown in Table~\ref{table:benchmarks_stats_human} and Table~\ref{table:benchmarks_stats_gpt4v}. Note that while the datasets in the VQA and captioning benchmarks originally have ground-truth answers, the instruction following benchmarks inherently does not have a reference answer. Using the same augmentation process mentioned in Section~\ref{section:3.1}, we augment a reference answer and a fine-grained score rubric for each instance within the LLaVA-Bench, VisIT-Bench, and \textsc{Perception-Bench}. For the \textsc{Perception-Bench}, which is our held-out test set, we also generate new instructions. For the VQA and captioning benchmarks, we generate 5 score rubrics with the original ground-truth answer in consideration. The authors manually checked the quality of the added components. 

\subsection{Setups \& Metrics}
Our evaluation setup is divided into 2 parts.

\paragraph{Setup \#1 (Table~\ref{table:benchmarks_stats_human})} In Section~\ref{section:5.1}, we utilize 45 instances with instance-wise hand-crafted score rubrics (15 instances each for \textsc{LLaVA-Bench}, \textsc{VisIT-Bench}, and \textsc{Perception-Bench}). We ask 9 human annotators proficient in English to provide a scoring decision as \textsc{Prometheus-Vision}. Then, we measure the correlation of the scoring decision by employing \textbf{Pearson}, \textbf{Kendall-Tau}, and \textbf{Spearman} as our metrics. Next, we ask human annotators to compare 2 language feedbacks that are sampled from either \textsc{GPT-4}, \textsc{GPT-4V}, or \textsc{Prometheus-Vision} (13B) and choose which one is better. Then, we measure the \textbf{Pairwise Preference Win-rate} between the 3 candidates. Details of the annotation setting are explained in Appendix~\ref{section:C.2}.

\paragraph{Setup \#2 (Table~\ref{table:benchmarks_stats_gpt4v})} In Section~\ref{section:5.2}, we expand the number of instances and utilize 1,085 fine-grained score rubrics tied across 3,560 instances in total. In this setting, we prompt GPT-4V three times and compare the correlation of the scoring decision by also prompting evaluator VLMs and evaluator LMs three times. As Setup \#1, we use \textbf{Pearson}, \textbf{Kendall-Tau}, and \textbf{Spearman} as our metrics.

\begin{figure}[!t]
\centering
  \noindent\includegraphics[width=\linewidth]{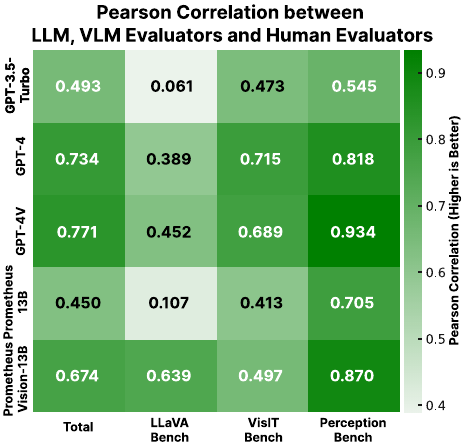} 
  \caption{Pearson Correlation between score decisions from human evaluators and score decisions from either \textsc{GPT-4V}, \textsc{GPT-4}, \textsc{GPT-3.5-Turbo}, \textsc{Prometheus-13B} and \textsc{Prometheus-Vision-13B} on 45 customized score rubrics from \textsc{LLaVA-Bench}, \textsc{VisIT-Bench}, and \textsc{Perception-Bench}. \textsc{Prometheus-Vision} shows a high correlation with human evaluators on instances with real-world images.}
  \label{fig:human_correlation}
\end{figure}

\section{Experimental Results}

\subsection{Can \textsc{Prometheus-Vision} Closely Simulate Human Evaluators?}\label{section:5.1} 
In this subsection, to verify whether \textsc{Prometheus-Vision} can emulate human evaluators, we measure the correlation between scores annotated by humans and those predicted by evaluator VLMs. The overall results are shown in Figure~\ref{fig:human_correlation}. 

\subsubsection{Correlation with Human Evaluators}
Our \textsc{Prometheus-Vision 13B} notably mirrors the high correlation exhibited by leading models \textsc{GPT-4} and \textsc{GPT-4V} on the \textsc{LLaVA-Bench} and \textsc{Perception-Bench}, achieving correlations of 0.639 and 0.870, respectively. Despite this, although our \textsc{Prometheus-Vision} outperforms \textsc{GPT-3.5-Turbo} and \textsc{Prometheus 13B} with a slightly improved correlation on the \textsc{VisIT-Bench}, it is lower than \textsc{GPT-4} and \textsc{GPT-4V}. 

We posit that this disparity primarily originates from the differing characteristics of the \textsc{VisIT-Bench} and other benchmarks. The former contains a higher proportion of text-rich images, such as graphs and charts, compared to the latter two datasets. Even though the \textsc{Perception Collection} also includes instruction sets for text-rich images, their amount is relatively limited. These inherent limitations in the model architecture of \textsc{Prometheus-Vision} present challenges in processing such text-rich images during inference. 

Nevertheless, recent works on vision-language models~\citep{zhang2023llavar,ye2023mplugowl2,kim2022donut,kim2023cream} show promising capabilities for handling these image types, providing a better backbone model for future iterations of \textsc{Prometheus-Vision}. In consideration of these findings, the use of text-rich datasets, along with the integration of new methods drawn from recent architectural advancements, could alleviate these limitations.

Also, it is worthwhile to compare where \textsc{GPT-4} (LM Evaluator) and \textsc{GPT-4V} (VLM Evaluator) excel at each benchmark. Similar to \textsc{Prometheus-Vision}, on the \textsc{VisIT-Bench}, \textsc{GPT-4} shows a slightly higher correlation with human evaluators compared to \textsc{GPT-4V}. This could mainly be because processing text is as important when assessing responses from text-rich images such as diagrams, charts, and graphs. On the other hand, \textsc{GPT-4V} shows a higher correlation with human evaluators on the \textsc{LLaVA-Bench} and \textsc{Perception-Bench} which includes diverse real-world images. 

\begin{figure}[!t]
\centering
  \noindent\includegraphics[width=\linewidth]{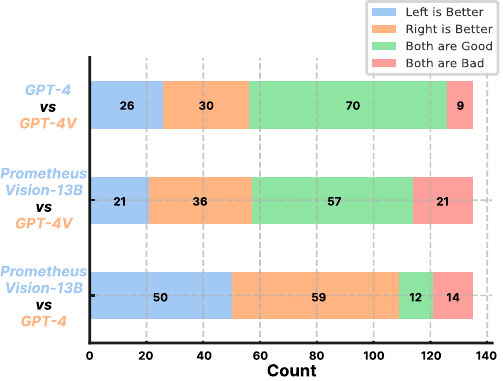} 
  \caption{Pairwise comparison of the quality of the language feedback generated by \textsc{GPT-4V}, \textsc{GPT-4}, and \textsc{Prometheus-Vision-13B}. Results show that \textsc{Prometheus-Vision}'s feedback is as good as or better than \textsc{GPT-4V}'s feedback 57.78\% of the time.}
  \label{fig:feedback_winrate}
\end{figure}



\subsubsection{Comparison of the Quality of the Feedback}
Next, we compare the quality of the language feedback generated by \textsc{GPT-4}, \textsc{GPT-4V}, and \textsc{Prometheus-Vision 13B} across 135 instances by hiring 9 human annotators. The detailed experimental setting is explained in Appendix~\ref{section:D} and the results are shown in Figure~\ref{fig:feedback_winrate}.

Surprisingly, The \textsc{Prometheus-Vision 13B} model is capable of generating feedback of a quality comparable to GPT-4. Among the 135 instances, human annotators determine that 57.78 \% of the time, \textsc{Prometheus-Vision}'s feedback is better or as good as \textsc{GPT-4V}'s feedback. Also, human annotators determine that 45.93 \% of the time, \textsc{Prometheus-Vision}'s feedback is better or as good as \textsc{GPT-4}'s feedback. These results indicate that \textsc{Prometheus-Vision} could also be utilized as a open-source critique model for assisting assessment by humans~\citep{saunders2022self}.

\begin{table*}[!ht]
\fontsize{10}{14}\selectfont
\centering
\resizebox{\textwidth}{!}{\begin{tabular}{lccccccccc}
    \toprule
    \multicolumn{1}{c}{\multirow{2}{*}{\textbf{Evaluator LM}}}& \multicolumn{3}{c}{\textsc{LLaVA-Bench}} & \multicolumn{3}{c}{\textsc{VisIT-Bench}} & \multicolumn{3}{c}{\textsc{Perception-Bench}}\\ 
    \cmidrule(lr){2-4} \cmidrule(lr){5-7} \cmidrule(lr){8-10} & Pearson & Kendall-Tau & Spearman & Pearson & Kendall-Tau & Spearman & Pearson & Kendall-Tau & Spearman\\ 
    \midrule
    \textsc{LLaVA-RLHF 7B}&0.328&0.379&0.412&0.317&0.193&0.215&0.415&0.337&0.374\\
    \textsc{LLaVA-RLHF 13B}&0.296&0.238&0.246&0.384&0.166 &0.185 &0.335&0.162&0.174\\ 
    \textsc{LLaVA-1.5 7B}&0.278&0.226&0.254&0.408 &0.188&0.214 & 0.602 & 0.383 & 0.419\\
    \textsc{LLaVA-1.5 13B}&-0.005&0.097&0.105&0.597&0.347 &0.376 &0.505&0.254&0.270\\
    \textsc{ShareGPT4V 7B}&0.366&0.222&0.247&0.360&0.222&0.256&0.474&0.338&0.378\\
    \textsc{Fuyu 8B}&-0.023&0.049&0.052&0.059&0.079&0.087&0.011&-2.15E-04&4.29E-06\\ \midrule
    \textsc{GPT-3.5-Turbo-0613}&0.107&0.221&0.243&\textbf{0.685}&\textbf{0.539}&\textbf{0.592}&0.563&0.379&0.417\\
    \textsc{Prometheus 7B}&0.233&0.192&0.210&0.482&0.363&0.419&0.723&0.491&0.534\\
    \textsc{Prometheus 13B}&0.376&0.327&0.365&0.514&0.352&0.406&0.705&0.468&0.513\\
    \textsc{GPT-4-0613}&\underline{0.712}&\underline{0.500}&\underline{0.530}&0.494&0.352&0.394&\underline{0.808}&\underline{0.626}&\underline{0.661}\\ \midrule
    \textsc{Prometheus-Vision 7B}&0.411&0.214&0.233&\underline{0.662}&\underline{0.424}&\underline{0.478}&0.700&0.471&0.502\\
    \textsc{Prometheus-Vision 13B}&\textbf{0.786}&\textbf{0.630}&\textbf{0.660}&0.574&0.378&0.425&\textbf{0.832}&\textbf{0.655}&\textbf{0.690}\\ \midrule
    \textsc{GPT-4V-Preview}&0.769&0.636&0.669&0.824&0.718&0.761&0.870&0.699&0.727\\
    \bottomrule    
\end{tabular}}
\caption{Pearson, Kendall-Tau, Spearman correlation with scores sampled from \textsc{GPT-4V} across 3 inferences on visual instruction following benchmarks. Note that \textsc{GPT-4V} was sampled 6 times in total to measure self-consistency. The best comparable statistics are in \textbf{bold} and second best are \underline{underlined} among baselines. We include GPT-4V as reference to show its self-consistency when inferenced multiple times.}
\label{table:benchmark_correlation}
\end{table*}

\begin{table*}[!ht]
\fontsize{10}{14}\selectfont
\centering
\resizebox{\textwidth}{!}{\begin{tabular}{lccccccccc}
    \toprule
    \multicolumn{1}{c}{\multirow{2}{*}{\textbf{Evaluator LM}}}& \multicolumn{3}{c}{\textsc{OKVQA}} & \multicolumn{3}{c}{\textsc{VQA v2}} & \multicolumn{3}{c}{\textsc{TextVQA}}\\ 
    \cmidrule(lr){2-4} \cmidrule(lr){5-7} \cmidrule(lr){8-10} & Pearson & Kendall-Tau & Spearman & Pearson & Kendall-Tau & Spearman & Pearson & Kendall-Tau & Spearman\\ 
    \midrule
    \textsc{LLaVA-RLHF 7B}&0.562&0.330&0.368&0.111&0.061&0.074&0.208&0.163&0.187\\
    \textsc{LLaVA-RLHF 13B}&0.615&0.377&0.411&0.072&0.066 &0.079 &0.362&0.291&0.320\\ 
    \textsc{LLaVA-1.5 7B}&0.605&0.405&0.464&0.200 &0.134&0.152 & 0.290 & 0.201 & 0.247\\
    \textsc{LLaVA-1.5 13B}&0.548&0.373&0.404&0.346&0.286 &0.309 &0.409&0.352&0.408\\
    \textsc{ShareGPT4V 7B}&0.528&0.385&0.445&0.281&0.258&0.293&0.300&0.233&0.271\\
    \textsc{Fuyu 8B}&0.143&0.147&0.162&0.193&0.163&0.179&0.176&0.174&0.193\\ \midrule
    \textsc{GPT-3.5-Turbo-0613}&0.371&0.307&0.374&0.370&0.345&0.391&0.436&0.350&0.424\\
    \textsc{Prometheus 7B}&0.422&0.206&0.240&0.253&0.260&0.296&0.501&0.412&0.483\\
    \textsc{Prometheus 13B}&0.482&0.284&0.325&0.178&0.122&0.145&0.417&0.343&0.400\\
    \textsc{GPT-4-0613}&0.594&\textbf{0.509}&\textbf{0.584}&\textbf{0.605}&\textbf{0.527}&\textbf{0.606}&\textbf{0.723}&\textbf{0.642}&\textbf{0.718}\\ \midrule
    \textsc{Prometheus-Vision 7B}&\underline{0.608}&0.261&0.290&\underline{0.455}&\underline{0.395}&0.298&0.487&0.413&0.485\\
    \textsc{Prometheus-Vision 13B}&\textbf{0.653}&\underline{0.401}&\underline{0.441}&0.393&0.389&\underline{0.428}&\underline{0.512}&\underline{0.445}&\underline{0.523}\\ \midrule
    \textsc{GPT-4V-Preview}&0.795&0.735&0.810&0.681&0.610&0.684&0.791&0.705&0.796\\
    \bottomrule    
\end{tabular}}
\caption{Pearson, Kendall-Tau, Spearman correlation with scores sampled from \textsc{GPT-4V} across 3 inferences on visual question answering benchmarks. Note that \textsc{GPT-4V} was sampled 6 times in total to measure self-consistency. We include GPT-4V as reference to show its self-consistency when inferenced multiple times. For all questions, we provided the Evaluator VLM with a fine-grained rubrics.}
\label{table:vqa_correlation}
\end{table*}

\begin{table}
\centering
\fontsize{7.5}{9}\selectfont
\begin{tabular}{lcc}
    \toprule
    \multicolumn{1}{c}{\multirow{2}{*}{\textbf{Evaluator LM}}}& \textsc{COCO-Captions} & \textsc{No Caps}\\ 
    \cmidrule(lr){2-3} & Pearson & Pearson\\
    \midrule
    \textsc{LLaVA-RLHF 7B}&0.148&0.210\\
    \textsc{LLaVA-RLHF 13B}&0.198&0.171\\ 
    \textsc{LLaVA-1.5 7B}&0.248&0.155\\
    \textsc{LLaVA-1.5 13B}&0.157&0.111\\
    \textsc{ShareGPT4V 7B}&0.184&0.185\\
    \textsc{Fuyu 8B}&0.191&0.064\\ \midrule
    \textsc{GPT-3.5-Turbo-0613}&0.233&0.242\\
    \textsc{Prometheus 7B}&0.335&0.165\\
    \textsc{Prometheus 13B}&0.215&0.279\\
    \textsc{GPT-4-0613}&\underline{0.470}&\textbf{0.427}\\ \midrule
    \textsc{Prometheus-Vision 7B}&0.434&0.327\\
    \textsc{Prometheus-Vision 13B}&\textbf{0.508}&\underline{0.417}\\ \midrule
    \textsc{GPT-4V-Preview}&0.579&0.638\\
    \bottomrule    
\end{tabular}
\caption{Pearson, Kendall-Tau, Spearman correlation with scores sampled from \textsc{GPT-4V} across 3 inferences on captioning benchmarks. Note that \textsc{GPT-4V} was sampled 6 times in total to measure self-consistency. We include GPT-4V as reference to show its self-consistency when inferenced multiple times. For all questions, we provide the Evaluator VLM with a fine-grained rubrics.}
\label{table:captioning_correlation}
\vspace{-3mm}
\end{table}  

\subsection{Can \textsc{Prometheus-Vision} Closely Simulate GPT-4 Vision as a Judge?}\label{section:5.2}
In this subsection, to check whether \textsc{Prometheus-Vision} could be used as a reliable evaluator on various multi-modal tasks, we compare the correlation between scores predicted by GPT-4V and scores predicted by baselines including \textsc{Prometheus-Vision}. The results are shown in Tables~\ref{table:benchmark_correlation}, \ref{table:vqa_correlation}, \ref{table:captioning_correlation}.

\subsubsection{Visual Instruction Following Benchmarks}
The results in Table~\ref{table:benchmark_correlation} show that \textsc{Prometheus-Vision} demonstrates a higher correlation with GPT-4V compared to that of its backbone model, \textsc{LLaVA-v1.5}, in all 3 benchmarks and 2 model sizes. This indicates that training with \textsc{Perception Collection} enhances the VLM’s evaluation capabilities. Furthermore, in the \textsc{LLaVA-Bench} and \textsc{Perception-Bench}, \textsc{Prometheus-Vision 13B} exhibits a higher correlation than the LM evaluators GPT-3.5-Turbo and GPT-4.

\subsubsection{Visual Question Answering Benchmarks} 
Table~\ref{table:vqa_correlation} presents the correlation results in the visual question answering (VQA) benchmarks. In this benchmark, \textsc{Prometheus-Vision} significantly outperforms other open-source models, including \textsc{LLaVA-v1.5}. Also, we observe that \textsc{Prometheus-Vision}'s correlation is generally lower in VQA benchmarks compared to visual instruction following benchmarks. We attribute this to the \textsc{Perception Collection} training data, which generally involves longer responses, while the answers in the VQA benchmark are mostly short. Future works could consider adding more diversity to the training data to obtain a stronger VLM evaluator.

\subsubsection{Captioning Benchmarks} 
Unlike visual instruction following or VQA benchmarks, captioning benchmarks do not have a direct question but rather require writing a description of a given image in a short sentence. Therefore, we created prompts such as `Generate a coco-style caption.' and fed them to our evaluator VLM baselines during experiments. The results are shown in Table~\ref{table:captioning_correlation}. While most evaluators, including proprietary LMs, show low correlation, \textsc{Prometheus-Vision 13B} surprisingly stands out by showing a correlation above 0.5 in the COCO-Captions, indicating it could generalize to evaluate other visual-language tasks beyond its training data.

\begin{figure}
    \centering
    \includegraphics[width=1\linewidth]{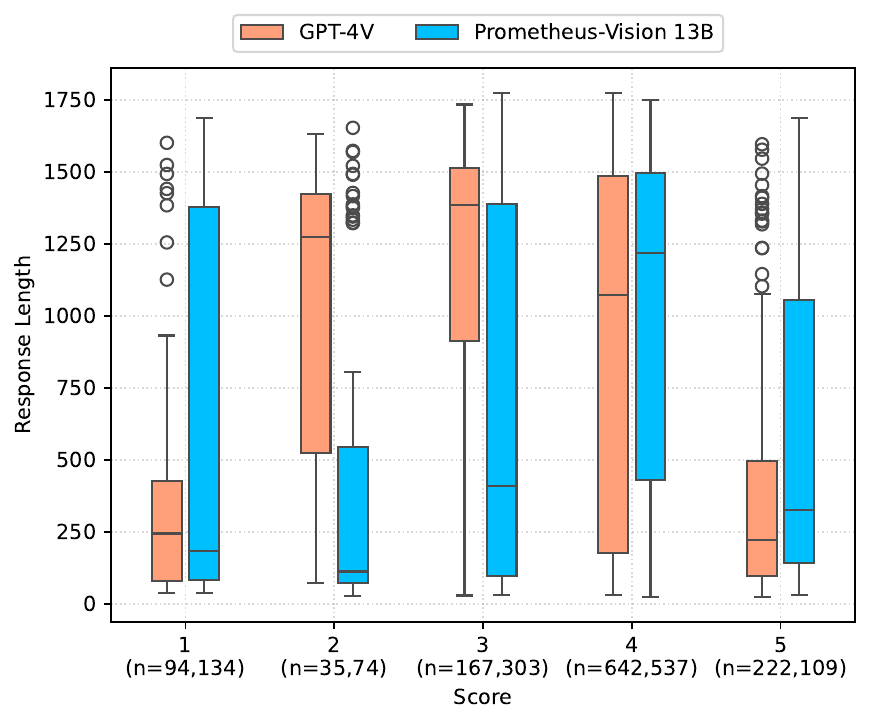}
    \caption{Distribution of length of responses by GPT-4V across different scores, as evaluated by GPT-4V and \textsc{Prometheus-Vision} 13B, in all test sets. Each score category on the x-axis is annotated with the quantity of responses that received that particular score from each Evaluator VLM. Individual test set results are in Figure~\ref{fig:response_length_test_individual}.}
    \label{fig:response_length_test_all}
\end{figure}

\section{Analysis of Potential Biases from VLM Evaluators}

\subsection{Is there a Length Bias?}\label{section:6.5}

Previous works have highlighted a phenomenon known as \textit{length bias} in models, which refers to a tendency of evaluator models to prefer longer responses~\citep{alpaca_eval, dubois2023alpacafarm, zheng2023judging}. This is a critical factor to consider during evaluation, as evaluators with length bias could give higher scores simply based on the length of the response, regardless of its actual content. To verify if this is the case, we plot and analyze the lengths of responses using our results from Section~\ref{section:5.1}.


The box plot in Figure~\ref{fig:response_length_test_all} showcases GPT-4V and \textsc{Prometheus-Vision} do not indiscriminately favor longer answers, indicating an absence of length bias. This is likely because our experimental setting is in an absolute grading setting where the evaluator VLM assesses the given responses with an absolute score rather than comparing two responses. This also aligns with the previous finding from \citet{zheng2023judging} and \citet{kim2023prometheus}. We provide more details of our analysis in Appendix~\ref{section:A.4} and Appendix~\ref{section:D}.

\begin{figure}[!t]
    \centering
    \begin{subfigure}{\linewidth}
        \includegraphics[width=\linewidth]{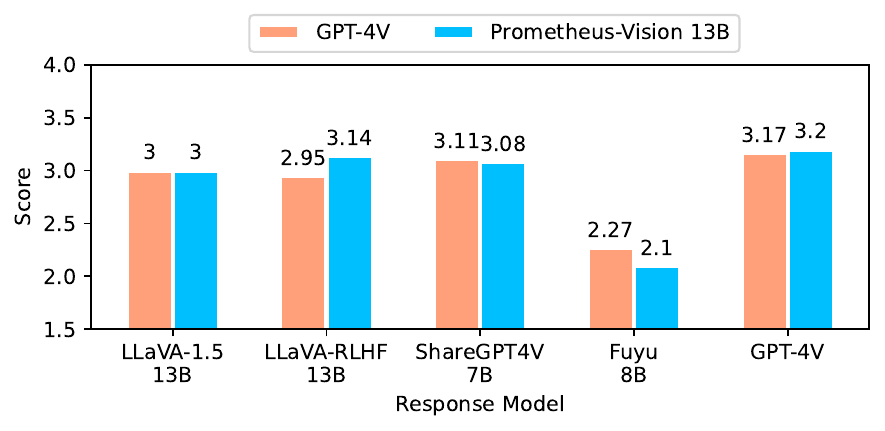}
        \caption{LLaVA-Bench}
    \end{subfigure}


    \begin{subfigure}{\linewidth}
        \includegraphics[width=\linewidth]{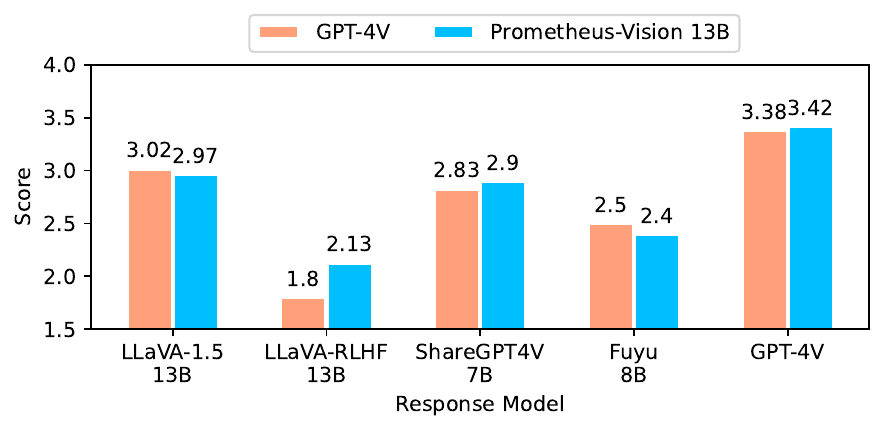}
        \caption{Perception-Bench}
    \end{subfigure}
    
    \caption{Evaluation of 5 VLMs on (a) LLaVA-Bench and (b) \textsc{Perception-Bench} using either \textsc{Prometheus-Vision} or \textsc{GPT-4V} as an evaluator VLM. Trends show that \textsc{Prometheus-Vision} could closely simulate \textsc{GPT-4V} evaluation. In addition, the open-source nature of \textsc{Prometheus-Vision} provides accessible and transparent evaluation for those developing State-of-the-Art VLMs.}
    \label{fig:model_score_comparison}
\end{figure}

\subsection{Is there a Self-Enhancement Bias?}\label{section:6.4}
Self-enhancement bias is another type of well-known bias where evaluators tend to prefer its own response~\citep{zheng2023judging}. Since \textsc{Prometheus-Vision} is a model specialized for evaluation purposes only, it does not directly suffer from this bias. However, since we train \textsc{Prometheus-Vision} with data augmented from GPT-4V and use LLaVA-v1.5 as our base model, this could indirectly influence \textsc{Prometheus-Vision}, making things more complicated. To investigate whether there is a self-enhancement bias, we analyze the trends of which score was given to different response VLMs on the LLaVA-Bench and Perception Bench. 

Figure~\ref{fig:model_score_comparison} illustrates the results. Overall, the results show that \textsc{Prometheus-Vision} and GPT-4V exhibit similar evaluation patterns across the two benchmarks, reinforcing the findings from previous correlation studies with GPT-4V. Notably, \textsc{Prometheus-Vision} gives a higher score to other models compared to its backbone model (LLaVA-v1.5) on the LLaVA-Bench, indicating that evaluator VLMs might not always prefer the responses from its backbone model.

While \textsc{Prometheus-Vision} does give the highest score to GPT-4V, it is hard to determine if this is because \textsc{Prometheus-Vision} was trained on data augmented from GPT-4V, or GPT-4V is distinctively better than the open-source VLMs. We leave analysis of this to future research.

Lastly, the trends from Figure~\ref{fig:model_score_comparison} also highlight the potential of our held-out testset, the \textsc{Perception-Bench}, to be used as a testbed for VLM development in future research. Specifically, on the predominant LLaVA-Bench, LLaVA-RLHF shows only a marginal difference of 0.14 points from GPT-4V. However, this gap widens significantly to 1.43 in \textsc{Perception Bench}. Since the \textsc{Perception Bench} was generated based on fine-grained rubrics, its instructions are more complex and extended responses than those of LLaVA-Bench.


\section{Conclusion}

In this paper, we expand the `LM-as-a-Judge' paradigm to the multi-modal space and introduce `VLM-as-a-Judge'. We first propose a multi-modal feedback dataset called the \textsc{Perception Collection}, which has unique score criteria for each instance, unlike the existing multi-modal datasets that do not heavily consider the important values to consider during evaluation. Using the \textsc{Perception Collection}, we train \textsc{Prometheus-Vision}, an open-source model specialized for evaluation purposes. The uniqueness of \textsc{Prometheus-Vision} is that it could adhere to user-defined criteria during evaluation. Through experiments, we show that \textsc{Prometheus-Vision} paves way for accessible and transparent evaluation of VLMs. We hope our work could pave the way for more research on open-source evaluators in different modalities.

\section*{Limitations}
One limitation of \textsc{Prometheus-Vision} is that it does not show optimal performance when evaluating instances that include text-rich images including diagrams, charts, and graphs. This is heavily reliant on the performance of the visual encoder used during visual instruction tuning of the backbone model, LLaVA-v1.5~\citep{liu2023visual,liu2023improved}. In the future, better VLM backbones could possibly resolve this issue. Moreover, another reason might come from the fact that the \textsc{Perception Collection} is heavily skewed towards real-world images, not text-rich images. Adding more feedback data that includes text-rich images could be an interesting line of future work.

Also, our work does not consider cases when images generated by image generation models are given as input. Future work could consider exploring whether VLM evaluators could assess text outputs conditioned on AI-generated images.

Lastly, as mentioned in our motivation for creating the \textsc{Perception Collection}, currently there are not a lot of multi-modal feedback datasets available for public use, compared to the text-only domain. Investigation of different forms of feedback, preference, and critique datasets would be an interesting line of future work. 


\bibliography{anthology,custom}

\appendix
\clearpage

\section{Analysis of \textsc{Perception Collection}}\label{section:A}
\subsection{Diversity of Score Rubrics}\label{section:A.1}

\begin{figure}
    \centering
    \includegraphics[width=1\linewidth]{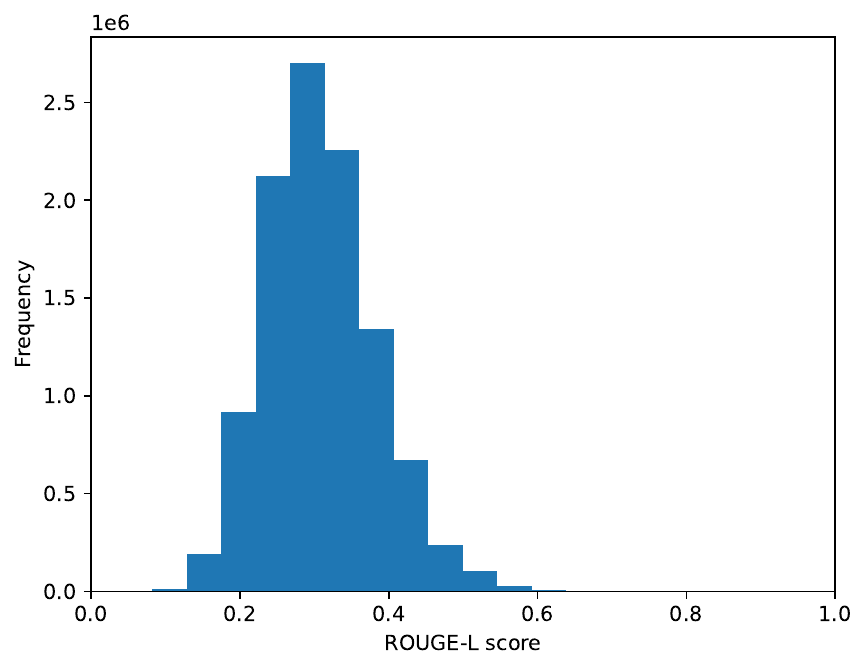}
    \caption{Distribution of ROUGE-L similarities in pairs of score rubric descriptions within \textsc{Perception-Collection}.}
    \label{fig:rouge_train}
\end{figure}
\begin{figure*}
    \centering
    \includegraphics[width=1\linewidth]{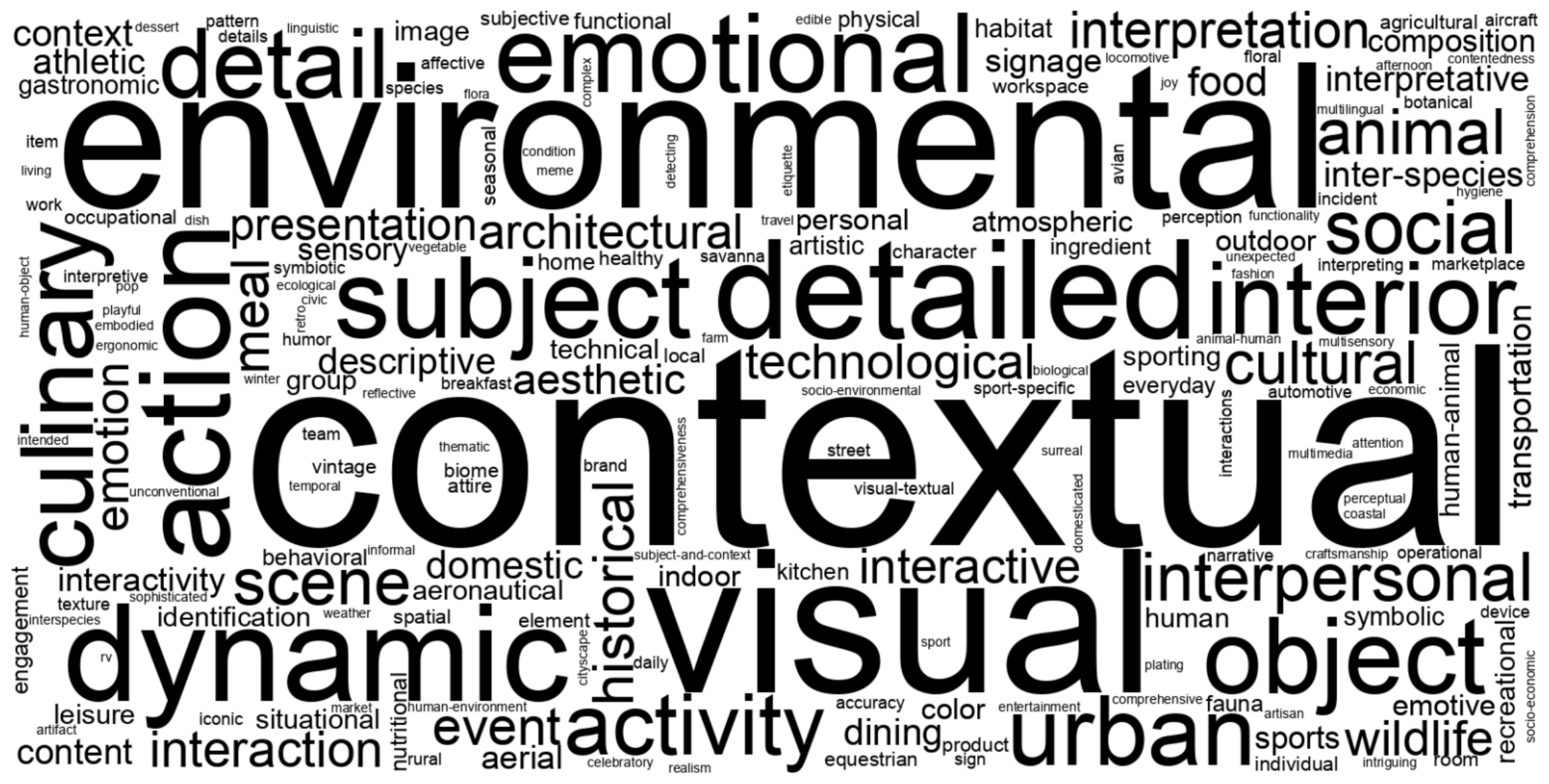}
    \caption{Word cloud of keywords in general-purpose score rubrics within \textsc{Perception-Collection}}
    \label{fig:general_wordcloud}
\end{figure*}
\begin{figure*}
    \centering
    \includegraphics[width=1\linewidth]{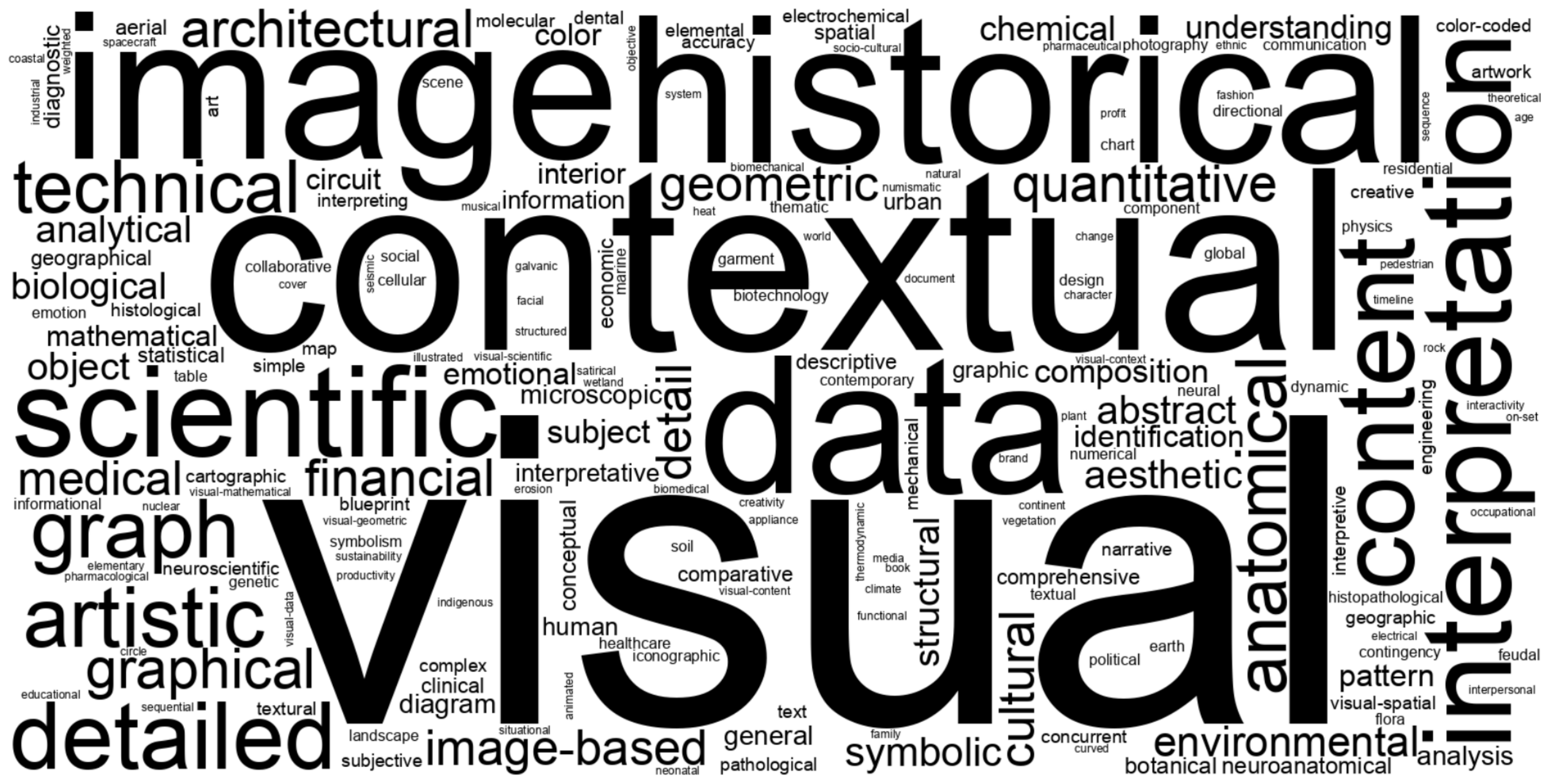}
    \caption{Word cloud of keywords in domain-specific score rubrics within \textsc{Perception-Collection}}
    \label{fig:specific_wordcloud}
\end{figure*}

When hand-crafting seed rubrics and generating new fine-grained score rubrics through brainstorming, for each rubric, we tag keywords that best describe the criteria. Figure~\ref{fig:general_wordcloud} and Figure~\ref{fig:specific_wordcloud} show word clouds of keywords in general-purpose rubrics and domain-specific rubrics included in \textsc{Perception-Collection}, respectively. General-purpose rubrics encourage a broader, more holistic perspective into the image as noted by the prominence of the words `environmental', `scene', `social', \textit{etc.}. Domain-specific rubrics bring more attention to the visual aspects of the image and data, specifying long-tail subfields of various subjects which are shown by the words `scientific', `artistic',`anatomical', \textit{etc.}. \\
Following previous works on machine-generated instructions \citep{wang-etal-2023-self-instruct, honovich-etal-2023-unnatural, kim2023prometheus}, we quantify the overlap of the generated score rubrics in our training data. Specifically, we compute ROUGE-L similarities between score rubric descriptions for every possible pair within \textsc{Perception-Collection}. The ROUGE-L distribution is plotted in Figure~\ref{fig:rouge_train}, with the average ROUGE-L score being 0.31 and the distribution being left-skewed. This low similarity score underscores the unique and varied nature of the \textsc{Perception Collection}.

\subsection{Decisiveness of Score Descriptions}\label{section:A.3}
\begin{figure}
    \centering
    \includegraphics[width=1\linewidth]{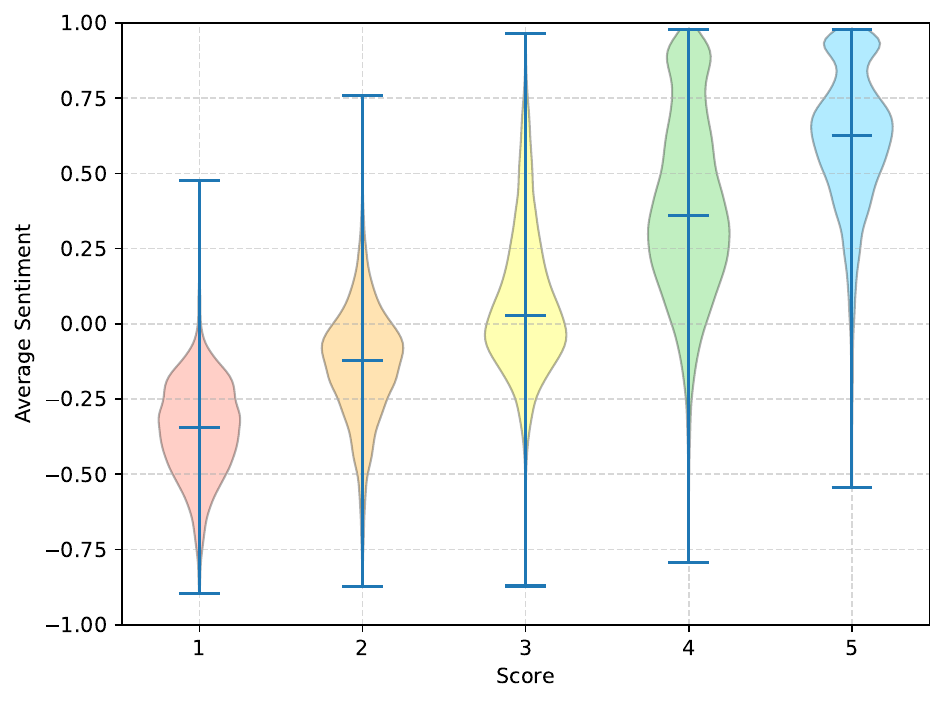}
    \caption{Average sentiment of descriptions for each score in \textsc{Perception-Collection} Rubrics. Sentiment of +1 signifies positivity, 0 neutrality, and -1 negativity.}
    \label{fig:sentiment_score}
\end{figure}
We examine whether each level of the scoring system in the rubric is clear and distinct. Following \citet{kim2023prometheus}, we compute the average sentiment in the description of each score in rubrics within \textsc{Perception-Collection}. We use a publicly available DeBERTa-distilled DistilBERT for sentiment analysis tasks~\citep{lik_xun_yuan_2023}. The results can be found in Figure~\ref{fig:sentiment_score}, where descriptions corresponding to a score of 1 are generally more negative, while those with a score of 5 are more positive. This suggests that the training data is appropriately interpolated according to scores and \textsc{Prometheus-Vision} trained on this dataset can conduct absolute scoring clearly and effectively. 

\subsection{Length Bias of Responses per Score Provided for Training}\label{section:A.4}
\begin{figure}
    \centering
    \includegraphics[width=1\linewidth]{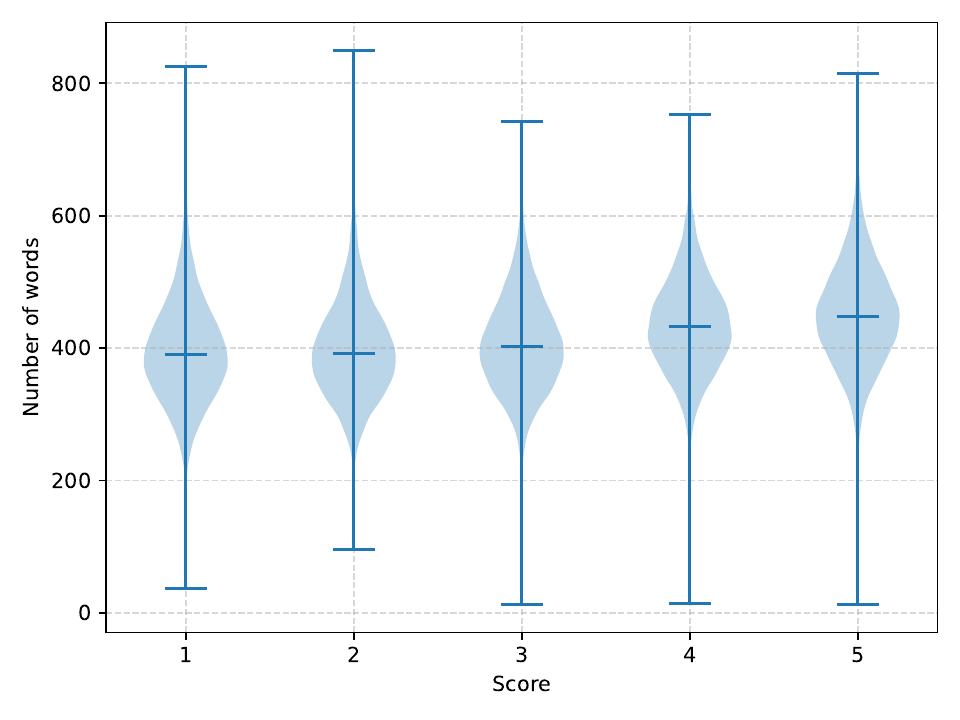}
    \caption{Distribution of length of responses scoring from 1 to 5 provided for training.}
    \label{fig:response_lenth_train}
\end{figure}
As explained in Section~\ref{section:3.1}, given an instruction, rubric, and reference answer, a response corresponding to score $i$ is generated for all $1 \le i \le 5$ to provide an evaluator model under training responses to practice assessment on. To nullify the tendency of recent LMs to give higher scores to longer responses \citep{alpaca_eval, dubois2023alpacafarm, zheng2023judging}, during \textsc{Perception-Collection} construction, we aim to maintain similar length of responses across the score range (See Appendix~\ref{section:E.1} for the exact prompt). The distribution of length of responses by score is plotted in Figure~\ref{fig:response_lenth_train}. Response lengths are distributed evenly across the score range, with an 417 words in average.

\begin{figure}
    \centering
    \includegraphics[width=1\linewidth]{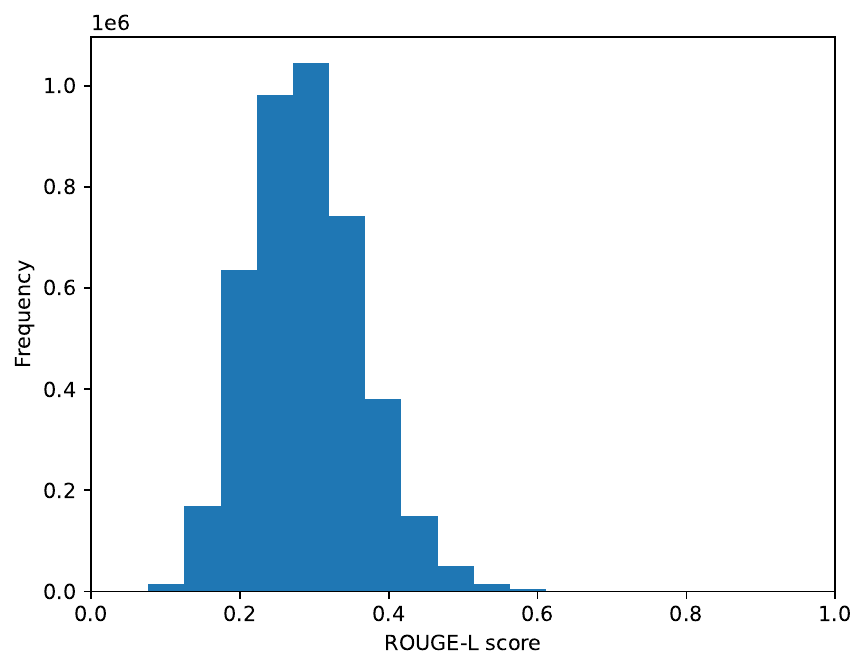}
    \caption{Distribution of ROUGE-L scores between score rubric descriptions in \textsc{Perception-Bench} and \textsc{Perception-Collection}.}
    \label{fig:rouge_test}
\end{figure}

\section{Analysis of \textsc{Perception-Bench}}\label{section:B}
\subsection{Validity of Unseen Score Rubrics}\label{section:B.1}
To ensure that \textsc{Perception-Bench} contains rubrics \textit{unseen} in \textsc{Perception-Collection}, we plot the ROUGE-L distribution between score rubric descriptions in \textsc{Perception-Bench} and \textsc{Perception-Collection} in Figure~\ref{fig:rouge_test}. The average ROUGE-L similarity between descriptions in our test set and train set is 0.29 and the distribution is left-skewed. We claim that the train-test overlap in our proposed dataset is low and that \textsc{Perception-Bench} contains many novel score rubrics.

\section{Comparison with conventional metrics}\label{section:6.2} 
Traditional VLM response evaluation metrics, which measure similarity solely between the reference answer and the response without considering the image, struggle to account for the varied information in images. Consequently, these conventional metrics can diverge significantly from human evaluations. As shown in Table~\ref{table:conventional_correlation}, there is a low Pearson correlation between human-predicted scores and conventional metrics. Notably, even METEOR, the conventional metric with the highest correlation, only achieves around 0.489, whereas \textsc{Prometheus-Vision 13B} demonstrates a higher correlation of 0.674. Moreover, conventional metrics often lack explainability. As Figure~\ref{fig:conventional_vlmjudge} indicates, they typically represent response quality with a simple value between 0 and 1. Model response, although it adequately depicts the image without employing expressions used in the reference answer, still receives a low score from conventional metrics due to their inability to perceive the image. In contrast, \textsc{Prometheus-Vision} not only provides a proper numeric score but also generates feedback that elucidates the reasons behind the score. This dual output can be instrumental in identifying ways to improve the model.
\begin{table}
\centering
\fontsize{7.5}{9}\selectfont
\begin{tabular}{lc}
    \toprule
    \multicolumn{1}{c}{\multirow{1}{*}{\textbf{Evaluator LM}}}& \textsc{LLaVA-VisIT-Perception} \\ 
    \cmidrule(lr){1-2} & Pearson \\
    \midrule
    \textsc{Rouge-1}&0.314\\
    \textsc{Rouge-L}&0.308\\
    \textsc{SPICE}&0.340\\
    \textsc{METEOR}&0.489\\ \midrule
    \textsc{GPT-3.5-Turbo}&0.493\\
    \textsc{Prometheus 13B}&0.450\\ 
    \textsc{GPT-4}&\textbf{0.734}\\ \midrule
    \textsc{Prometheus-Vision 13B}&\underline{0.674}\\ \midrule
    \textsc{GPT-4V-Preview}&0.771\\
    \bottomrule    
\end{tabular}
\caption{Pearson correlation with scores from human on 45 samples from 3 visual instruction following benchmarks. The best comparable statistics are \textbf{bolded} and second best \underline{underlined} among baselines.}
\label{table:conventional_correlation}
\vspace{-3mm}
\end{table}


\begin{figure*}[!ht]
    \centering
    \begin{subfigure}{0.3\linewidth}
        \includegraphics[width=\linewidth]{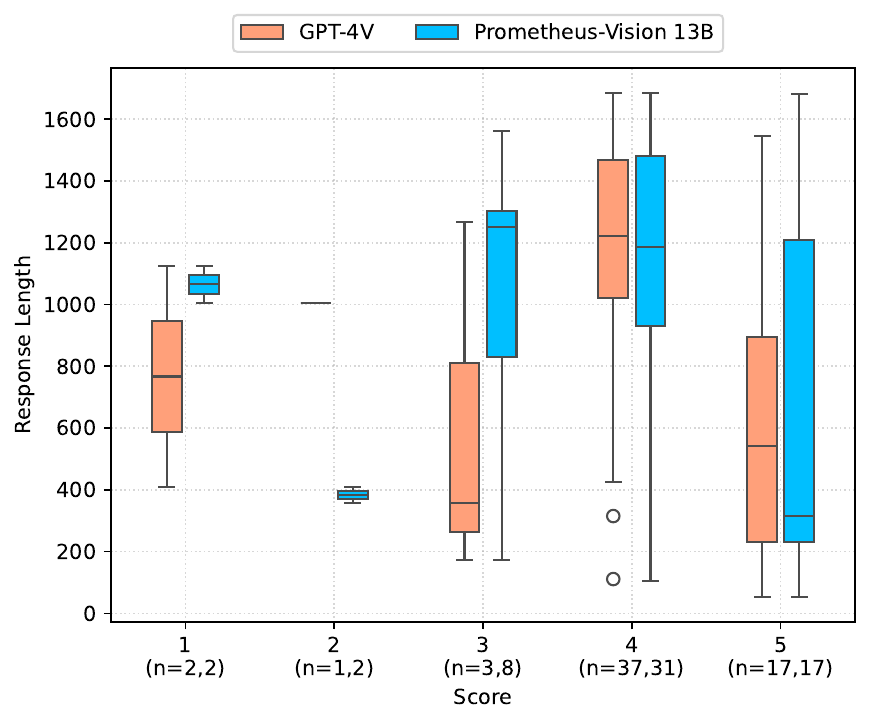}
        \caption{LLaVA-Bench}
    \end{subfigure}
    \hfill
    \begin{subfigure}{0.3\linewidth}
        \includegraphics[width=\linewidth]{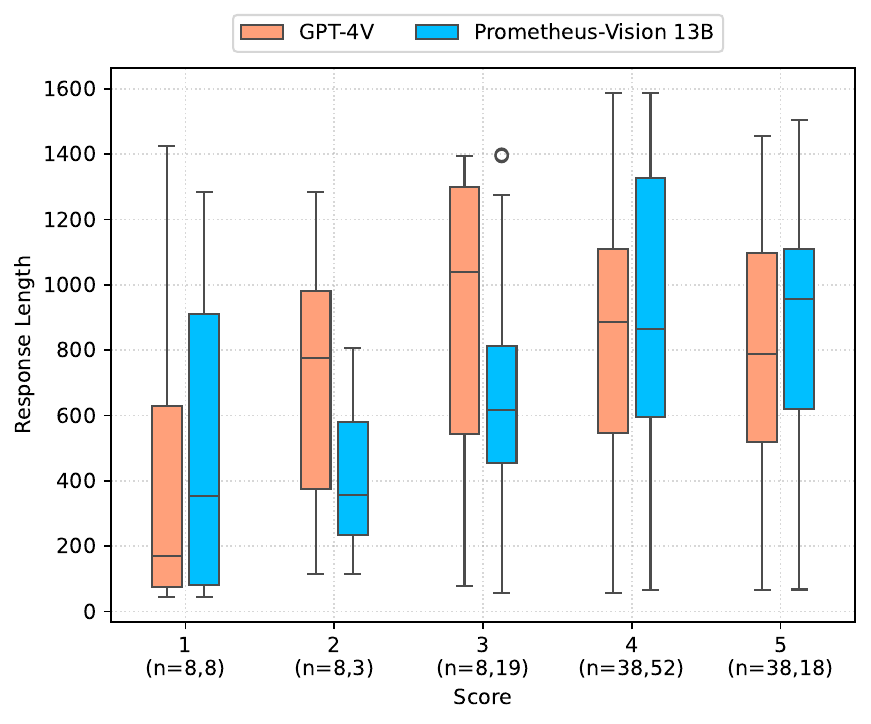}
        \caption{VisIT-Bench}
    \end{subfigure}
    \hfill
    \begin{subfigure}{0.3\linewidth}
        \includegraphics[width=\linewidth]{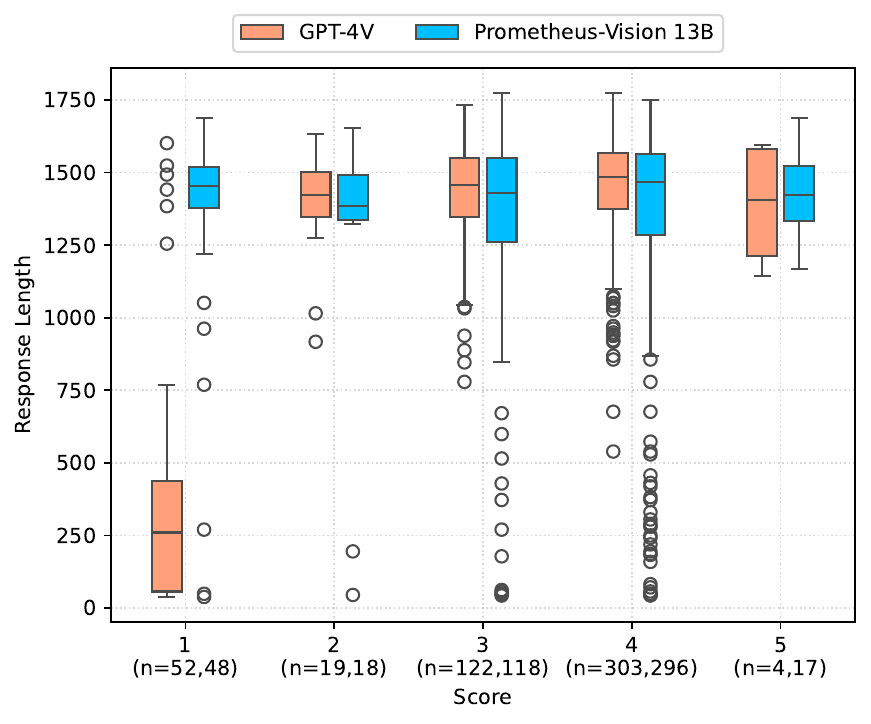}
        \caption{Perception-Bench}
    \end{subfigure}

    \medskip

    \begin{subfigure}{0.3\linewidth}
        \includegraphics[width=\linewidth]{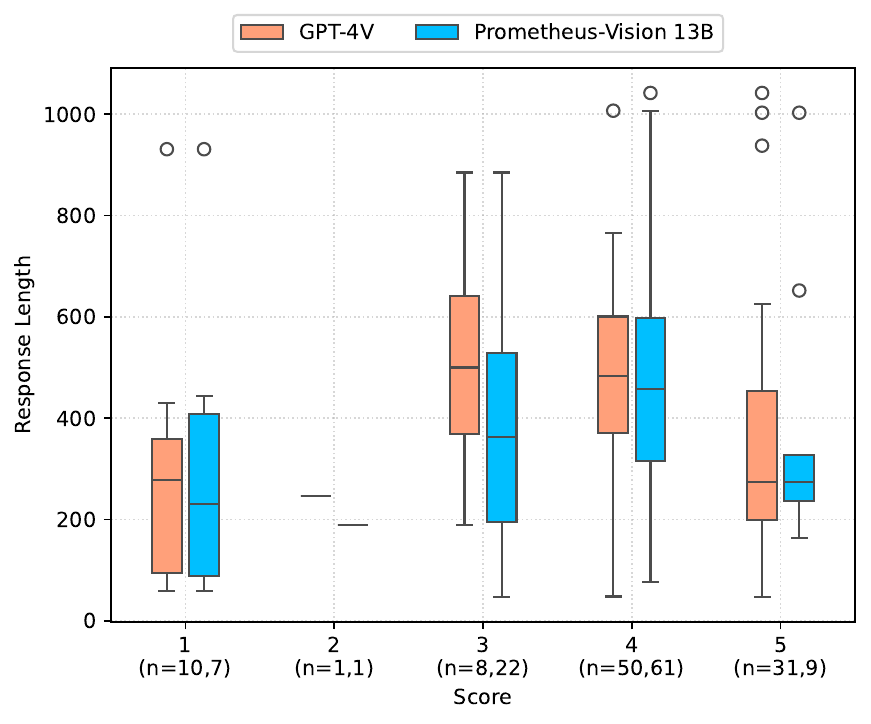}
        \caption{OKVQA}
    \end{subfigure}
    \hfill
    \begin{subfigure}{0.3\linewidth}
        \includegraphics[width=\linewidth]{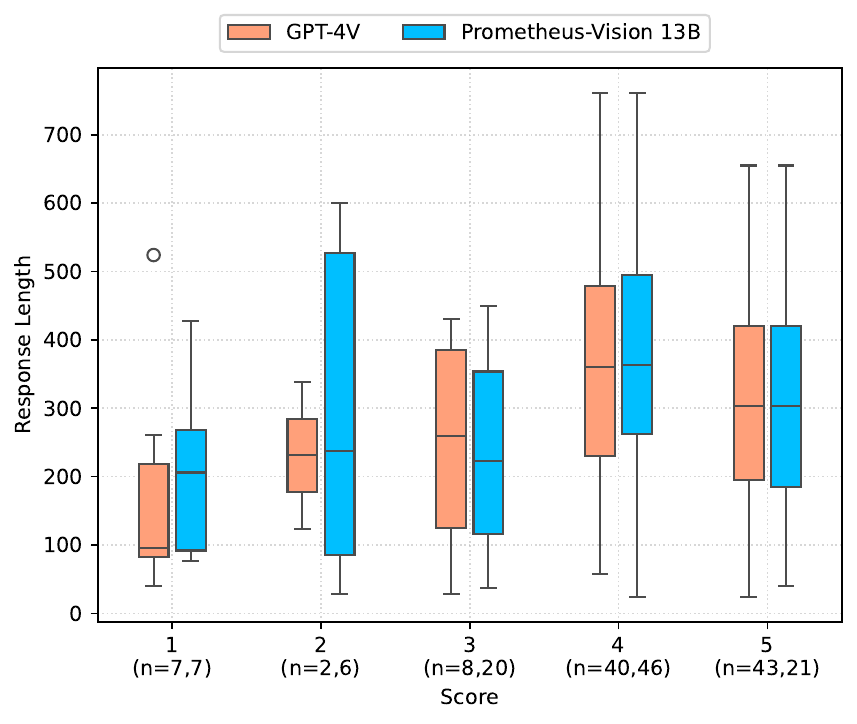}
        \caption{VQAv2}
    \end{subfigure}
    \hfill
    \begin{subfigure}{0.3\linewidth}
        \includegraphics[width=\linewidth]{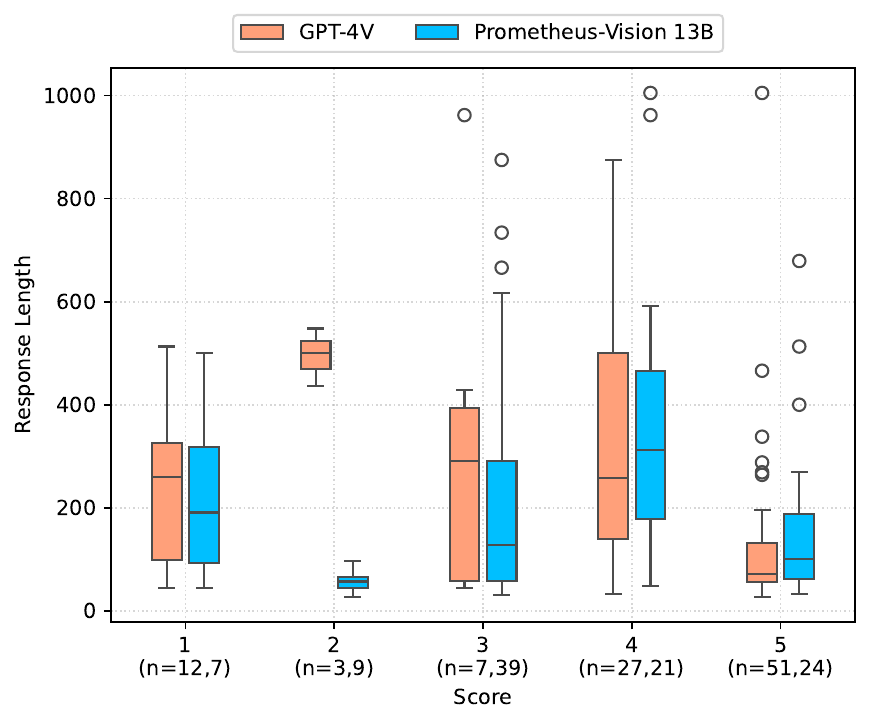}
        \caption{TextVQA}
    \end{subfigure}

    \medskip

    \begin{subfigure}{0.3\linewidth}
        \includegraphics[width=\linewidth]{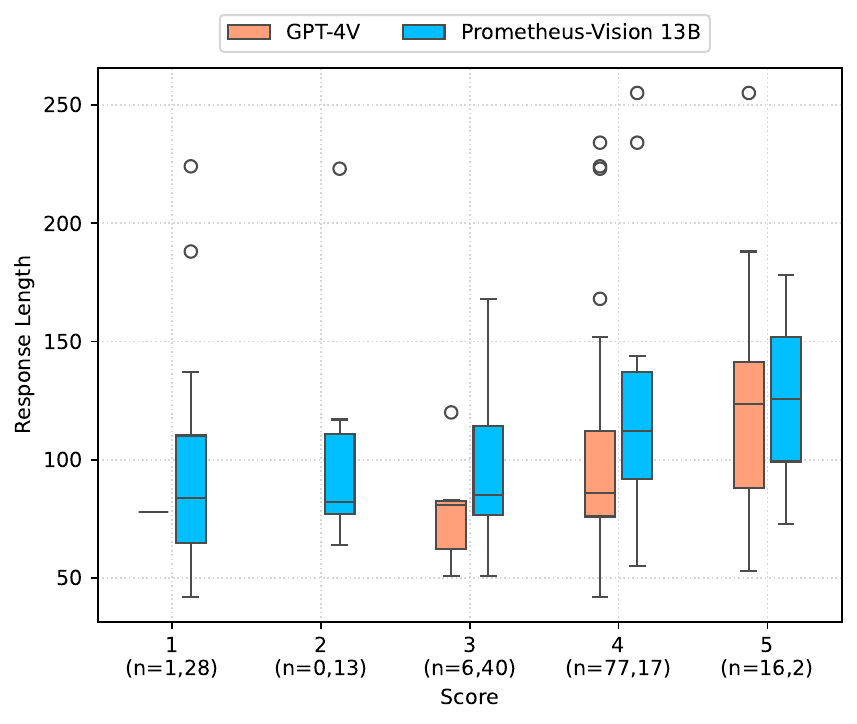}
        \caption{COCO-Captions}
    \end{subfigure}
    \hfill
    \begin{subfigure}{0.3\linewidth}
        \includegraphics[width=\linewidth]{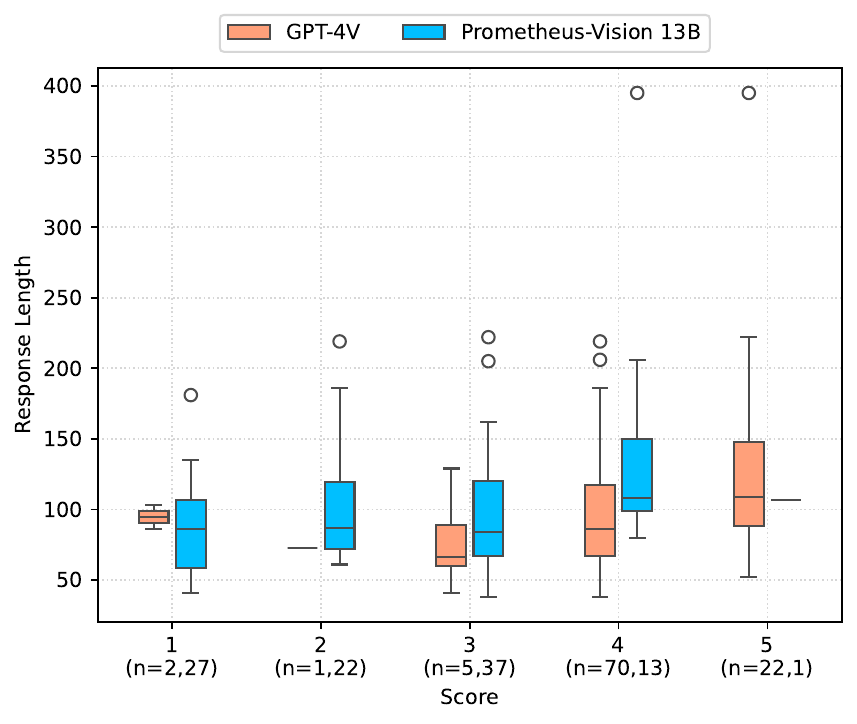}
        \caption{No Caps}
    \end{subfigure}

    \caption{Full distribution of length of responses by GPT-4V across different scores, as evaluated by GPT-4V and \textsc{Prometheus-Vision 13B}, in each test set. Each scoring category on the x-axis is annotated with the number of responses that received that particular score from each Evaluator VLM.}
    \label{fig:response_length_test_individual}
\end{figure*}

\section{Experimental Details}\label{section:C}
\subsection{Implementation Details and Computation}\label{section:C.1}
\textbf{Training} We employ LLaVA-1.5 7B / 13B as the backbone VLM for \textsc{Prometheus-Vision}. For the language model component, we utilize vicuna-13b-v1.5, and for the vision encoder, we use clip-vit-large-patch-14-336px. We freeze both the language model and the vision encoder, focusing our training solely on an MLP based alignment network. The training is conducted for one epoch, with a batch size per device set at 32. We set the learning rate at 1e-3, with no weight decay and a warmup ratio of 0.03. A cosine scheduler is utilized as the learning rate scheduler. To enhance training efficiency, we incorporate gradient checkpointing and deepspeed zero 2 in our training process. \\
\textbf{Inference} We use three Response VLMs to generate responses to given images and questions in each dataset. Then, an Evaluator VLM generates feedback and scores indicating how the response might improve given these responses, along with the image, question, reference answer, and a guiding rubric. This approach allows us to measure the correlation between scores from GPT-4V and those from other models. In the process of generating feedback, the model employs sampling with a temperature set to 1.0 and top-p set at 0.9, while the maximum number of tokens is configured to 2048. \\
Regarding the resources utilized for training and inference, the GPU setup includes 8 NVIDIA A100 80GB. For the CPU, an AMD EPYC 7543 32-Core Processor is used.

\subsection{Details in Human Evaluation}\label{section:C.2}
\begin{figure*}
    \centering
    \includegraphics[width=1\linewidth]{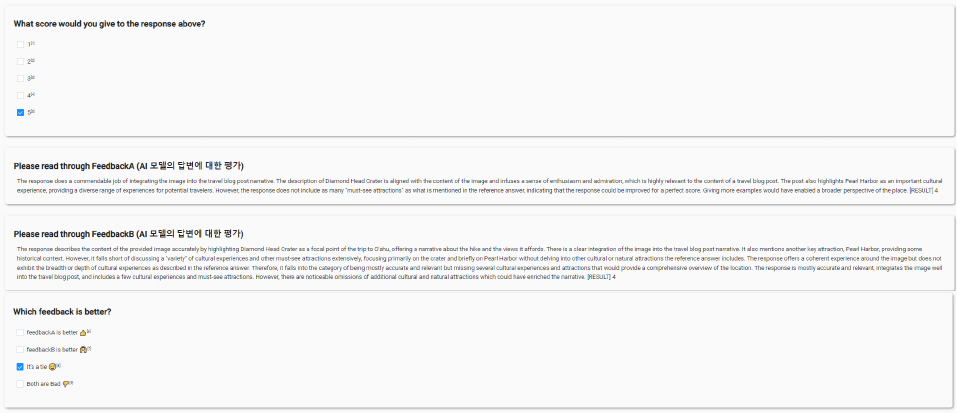}
    \caption{A screenshot of how human evaluators annotated their scoring decision and chose which feedback is better among different VLM, LM evaluator baselines.}
    \label{fig:annotation_screenshot2}
\end{figure*}

We recruit 9 undergraduate students proficient in English to conduct a human evaluation. The dataset used for the human evaluation is exclusively drawn from the Visual Instruction Tuning Benchmarks. Additionally, we randomly sample 15 items each from LLaVA-Bench, VisIT-Bench, and Perception-Bench, creating a total of 45 problems. For the pairwise feedback quality comparison, we utilize feedback from GPT-4V, GPT-4, and \textsc{Prometheus-Vision 13B}. Each of the 45 problems is structured to compare two out of the three feedbacks. Consequently, 3 sets of the same 45 problems are prepared, and the 9 participants are divided into 3 groups, with each group evaluating the same set of problems. We use Label Studio as the evaluation platform\footnote{\url{https://labelstud.io}}. The annotation interface is shown in Figure~\ref{fig:annotation_screenshot2}.

\section{Length Bias during Evaluation}\label{section:D}

We report GPT-4V response length distribution scored by GPT-4V and \textsc{Prometheus-Vision} 13B on individual test sets in Figure~\ref{fig:response_length_test_individual}. Overall trends show that both Evaluator VLMs do not display bias towards lengths in responses during inference.

\section{List of Prompts}\label{section:E}
\subsection{Prompts for \textsc{Perception Collection} Creation}\label{section:E.1}
We include the prompts used in the creation of our training dataset, \textsc{Perception Collection}. The Example Criteria include hand-crafted seed rubrics that were sampled and inserted beforehand. Additionally, for fine-grained rubric augmentation, the same prompt is used, but general-purpose rubrics and domain-specific rubrics are augmented separately, ensuring the seed rubrics are also individually incorporated without mixing. Notably, although the prompt does not feature an image insertion, in practice, images are included when calling the GPT-4V API. Detailed information is in the OpenAI API document\footnote{\url{https://platform.openai.com/docs/guides/vision}}.

\newpage
\begin{mybox}{Prompt for rubric augmentation}
You are helpful and creative rubric generator. You should brainstorm creative and impressive three rubrics used to evaluate the ability of a vision-language model to generate text when given an image.\\ \\
The rubric must be structured to assess areas that can be answered by viewing the image. It consists of a description explaining specific tasks and criteria for scoring. Here you will see 4 examples of 'criteria', and their scoring rubrics, formatted as JSON. \\ \\
Criteria 1: \\
\{Example Criteria 1\} \\ \\
Criteria 2: \\
\{Example Criteria 2\} \\ \\
Criteria 3: \\
\{Example Criteria 3\} \\ \\
Criteria 4: \\
\{Example Criteria 4\} \\ \\
Please brainstorm new three criterias and scoring rubrics. \\
Be creative and create new but useful criteria that people in different settings or industries might find practical. \\
Please format the output as same as the above examples with no extra or surrounding text. And you should not mention the term like `Criteria X:' and ```json'''. In JSON, all keys and string values must be enclosed in double quotes (""). For example, {"key": "value"} is a valid format, but {key: "value"} or {'key': 'value'} are not valid.\\
You should create a diverse rubrics suitable for the given image \\ \\
Generated criteria:
\end{mybox}
\begin{mybox}{Prompt for checking alignment}
You are helpful and creative rubric evaluator. You will be given one image and a rubric used to evaluate the capabilities of a vision-language model based on that image. If the rubric is well-aligned with the given image, you should answer 'align'. However, if the rubric does not fit the given image and there are areas for improvement, you should answer 'misalign'. \\ \\
The rubric must be structured to assess areas that can be answered by viewing the image. It consists of a description explaining specific tasks and criteria for scoring. Here you will see the rubric, and their scoring rubrics, formatted as JSON. \\ \\
Rubric: \\
\{Rubric\} \\ \\
Please answer 'align' or 'misalign'. You should generate the output in lowercase. \\ \\
Alignment:
\end{mybox}
\begin{mybox}{Prompt for refining rubric}
You are helpful and creative rubric creator. You will be given one image and a rubric used to evaluate the capabilities of a vision-language model based on that image. If the rubric does not fit the given image and there are areas for improvement, you should make improvements to create a better rubric. \\ \\
The rubric must be structured to assess areas that can be answered by viewing the image. It consists of a description explaining specific tasks and criteria for scoring. Here you will see the rubric, and their scoring rubrics, formatted as JSON. \\ \\
Rubric: \\
\{Rubric\} \\ \\
If there are areas that need improvement in the given rubric, improve the rubric that better fits the given image. Maximize your creativity to ensure that the rubric you refine is not too similar to the already existing one. \\ \\
Please format the output as same as the above examples with no extra or surrounding text. You should generate only one rubric. And you should not mention the term like `Criteria X:' and ```json'''. In JSON, all keys and string values must be enclosed in double quotes (""). For example, {"key": "value"} is a valid format, but {key: "value"} or {'key': 'value'} are not valid. \\ \\
Generated rubric:
\end{mybox}
\begin{mybox}{Prompt for generating instruction (1)}
Your job is to generate a new novel problem and a response that is related to the given score rubric and image. \\ \\
The score rubric: \\
\{Rubric\} \\ \\
* Problem \\
- The problem should inherently be related to the score criteria, score rubric and image given above. Specifically, the score criteria should be the core attributes required to solve the problem. \\
- The problem itself should not be too generic or easy to solve. \\
- Try to make the person who might solve the problem not notice the existence of the score rubric by not explicitly mentioning it, and also provide additional inputs and options if needed. \\
- Assume a situation where a user is interacting with an AI model. The user would try to ask in a first-person point of view, but not using terms like "I", "A User" or "You" in the first sentence. \\
- Do not give a role to the AI, assume that the user is asking a question from his point of view. \\
- Do not include any phrase related to AI model in the problem. \\
- The problem should only be answered by looking at an image, not just by reading the problem. \end{mybox}
\begin{mybox}{Prompt for generating instruction (2)}
* Response \\
- The response should be a response that would get a score of 5 from the score rubric. \\
- The response should be as detailed as possible unless the score rubric is related to conciseness or brevity. It should consist of multiple paragraphs, a list of items, or a step-by-step reasoning process. \\
- The response should look like how a well-prompted GPT-4 would normally answer your problem. \\ \\
* Format \\
- DO NOT WRITE ANY GREETING MESSAGES, just write the problem and response only. \\
- In front of the problem, append the phrase 'Problem:' and in front of the response, append the phrase 'Response:'. \\
- Write in the order of 'Problem' - 'Response', where the two items are separated by the phrase '[NEXT]'. \\
- Write [END] after you are done. \\ \\
Data Generation:
\end{mybox}
\begin{mybox}{Prompt for response and feedback (1)}
Your job is to generate a response that would get a score of \{score\} and corresponding feedback based on the given score rubric and image. For reference, a reference response that would get a score of 5 is also given.\\ \\
Instruction: \\
\{instruction\} \\ \\
The score rubric: \\
\{rubric\} \\ \\
Reference response (Score 5): \\
\{response\} \\ \\
* Response \\
- The quality of the score \{score\} response should be determined based on the score rubric and image, not by its length.\\
- The score \{score\} response should have the same length as the reference response, composed of \{number of sentences\} sentences.\\
- Do not explicitly state the keywords of the score rubric inside the response.\\ \\
\end{mybox}

\begin{mybox}{Prompt for response and feedback (2)}
* Feedback\\
- The score \{score\} feedback should each be an explanation of why the response would get a score of \{score\}. It should be written based on the generated response, score rubric and image.\\
- The score \{score\} feedback shouldn’t just copy and paste the score rubric, but it should also give very detailed feedback on the content of the corresponding response.\\
- The score \{score\} feedback should include the phrase 'So the overall score is \{score\}' in the last sentence.\\ \\
* Format\\
- DO NOT WRITE ANY GREETING MESSAGES, just write the problem and response only.\\
- In front of the response, append the phrase 'Response:' and in front of the feedback, append the phrase 'Feedback:'.\\
- Write in the order of 'Response' - 'Feedback', where the two items are separated by the phrase '[NEXT]'.\\
- Write [END] after you are done.\\ \\
Data Generation:
\end{mybox}

\subsection{Prompts for \textsc{Prometheus-Vision}}\label{section:E.2}
\begin{mybox}{Prompt for evaluation}
\#\#\#Task Description: \\
An instruction (might include an Input inside it), a response to evaluate, a reference answer that gets a score of 5, image and a score rubric representing an evaluation criterion is given. \\
1. Write a detailed feedback that assesses the quality of the response strictly based on the given score rubric, not evaluating in general. \\
2. After writing a feedback, write a score that is an integer between 1 and 5. You should refer to the score rubric. \\
3. The output format should look as follows: Feedback: (write a feedback for criteria) [RESULT] (an integer number between 1 and 5) \\
4. Please do not generate any other opening, closing, and explanations. \\ \\
\#\#\#The instruction to evaluate: \\
\{Instruction\} \\ \\
\#\#\#Response to evaluate: \\
\{Response\} \\ \\
\#\#\#Reference Answer (Score 5): \\
\{Reference answer\} \\ \\
\#\#\#Score Rubrics: \\
\{Description\} \\
Score 1: \{Criteria of score 1\} \\
Score 2: \{Criteria of score 2\} \\
Score 3: \{Criteria of score 3\} \\
Score 4: \{Criteria of score 4\} \\
Score 5: \{Criteria of score 5\} \\ \\
\#\#\#Feedback: \\
\end{mybox}
\newpage

\section{Qualitative Examples}\label{section:F}
\begin{figure*}
    \centering
    \includegraphics[width=1\linewidth]{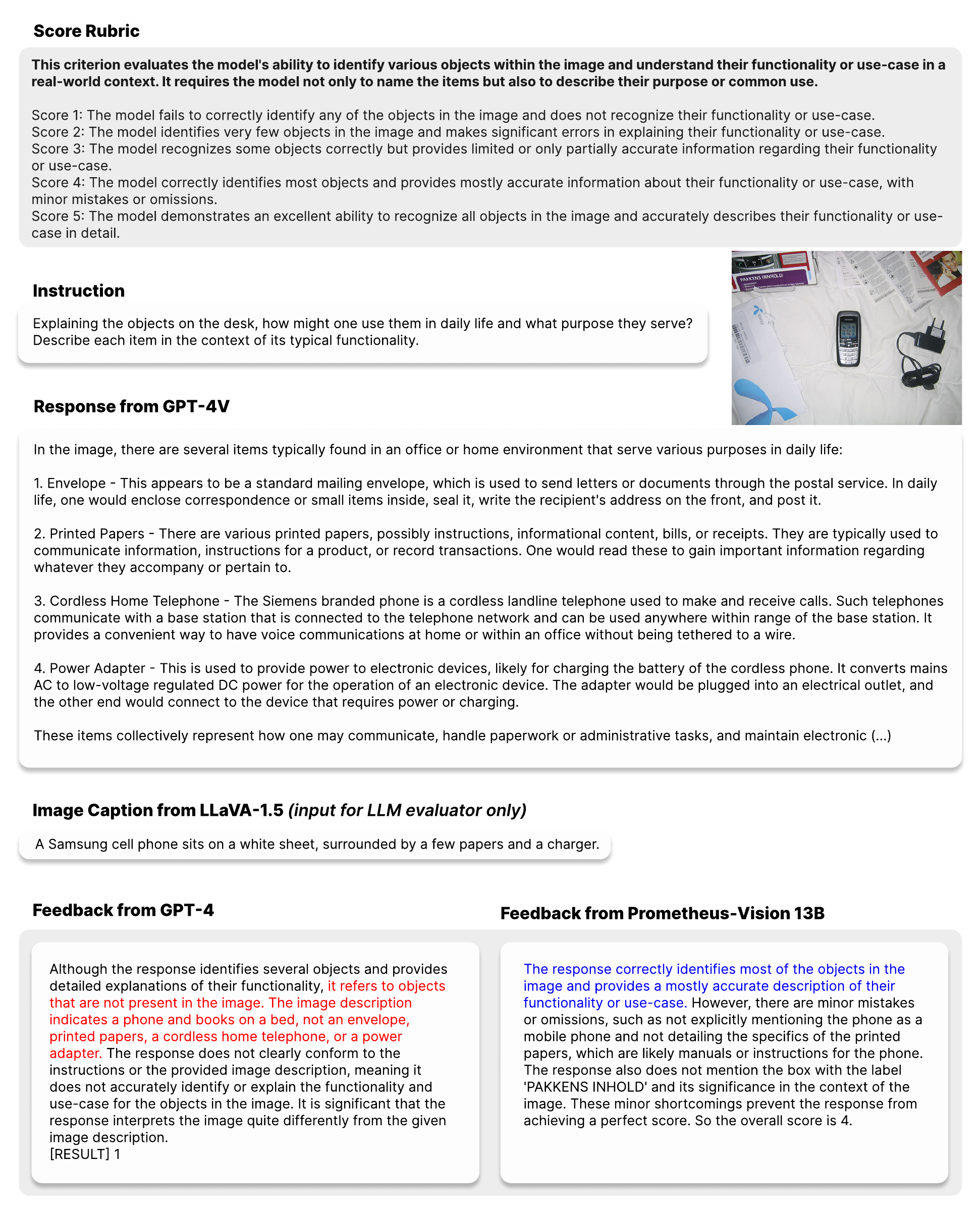}
    \caption{An example of comparison between GPT-4 feedback and \textsc{Prometheus-Vision} feedback on a \textsc{Perception-Bench} instance. In here, GPT-4 shows its limitation of evaluating a VLM output as an LM judge. As GPT-4 is not trained to process images, a description of the image produced by LLaVA-1.5 is provided to GPT-4 as a proxy for the actual image. GPT-4 cannot detect objects existing in the image that are \textit{not} mentioned in the image caption but are correctly mentioned in the response being evaluated. The incorrect parts of the feedback are in \textcolor{red}{red}, and for comparison, the correct parts are in \textcolor{blue}{blue}. Note that the reference answer is included in the input for both evaluators but is omitted in this figure for brevity.}
    \label{fig:cellphone_example}
\end{figure*}

\begin{figure*}
    \centering
    \includegraphics[width=1\linewidth]{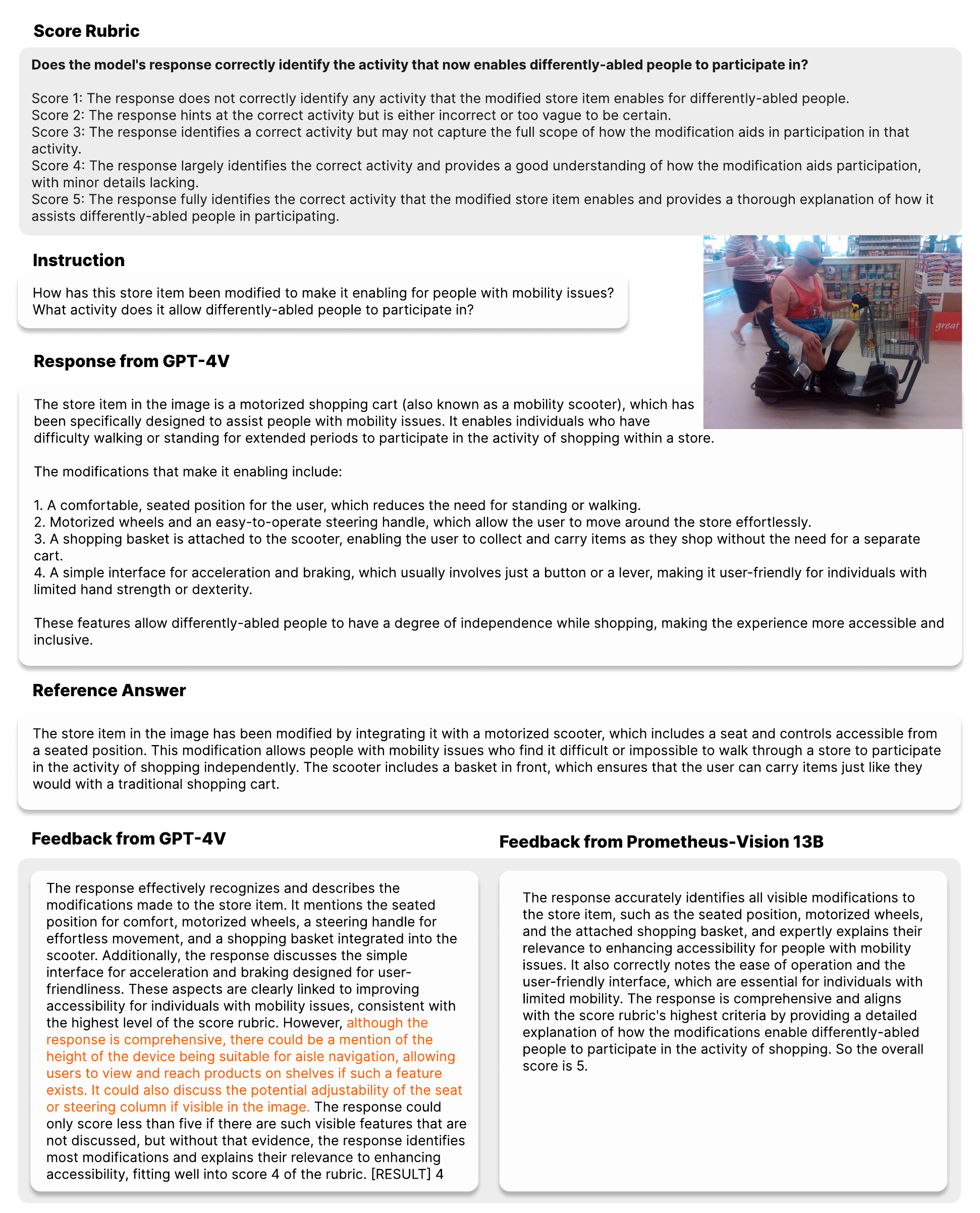}
    \caption{An example of comparison between GPT-4V feedback and \textsc{Prometheus-Vision} feedback on a VisIT-Bench instance. \textsc{Prometheus-Vision} follows the score rubric and reference answer well while GPT-4V gives an overly critical feedback (\textcolor{orange}{orange}).}
    \label{fig:mobility_example}
\end{figure*}

\begin{figure*}
    \centering
    \includegraphics[width=1\linewidth]{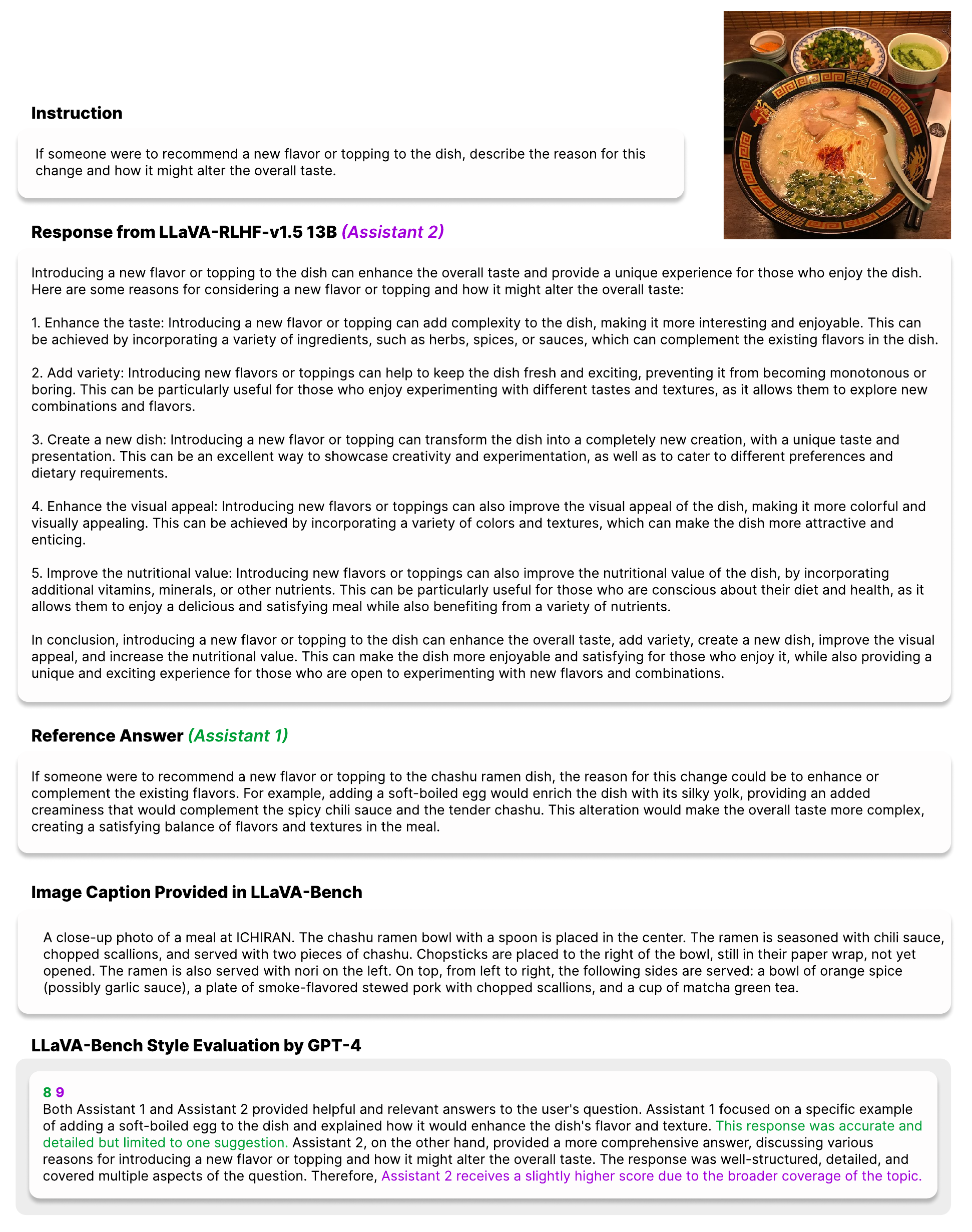}
    \caption{An example of LLaVA-Bench style GPT-4 evaluation of LLaVA-RLHF-v1.5 13B response on a LLaVA-Bench instance. The original LLaVA-Bench evaluation is performed by GPT-4 comparing the model response against the reference answer, where GPT-4 rates the reference answer (Assistant 1) first and then rates the model response (Assistant 2) second. LLaVA-RLHF-v1.5 13B gives a long, generic answer that can be generated without the image in context, yet is scored higher than the reference answer.}
    \label{fig:ramen_example}
\end{figure*}

\begin{figure*}
    \centering
    \includegraphics[width=1\linewidth]{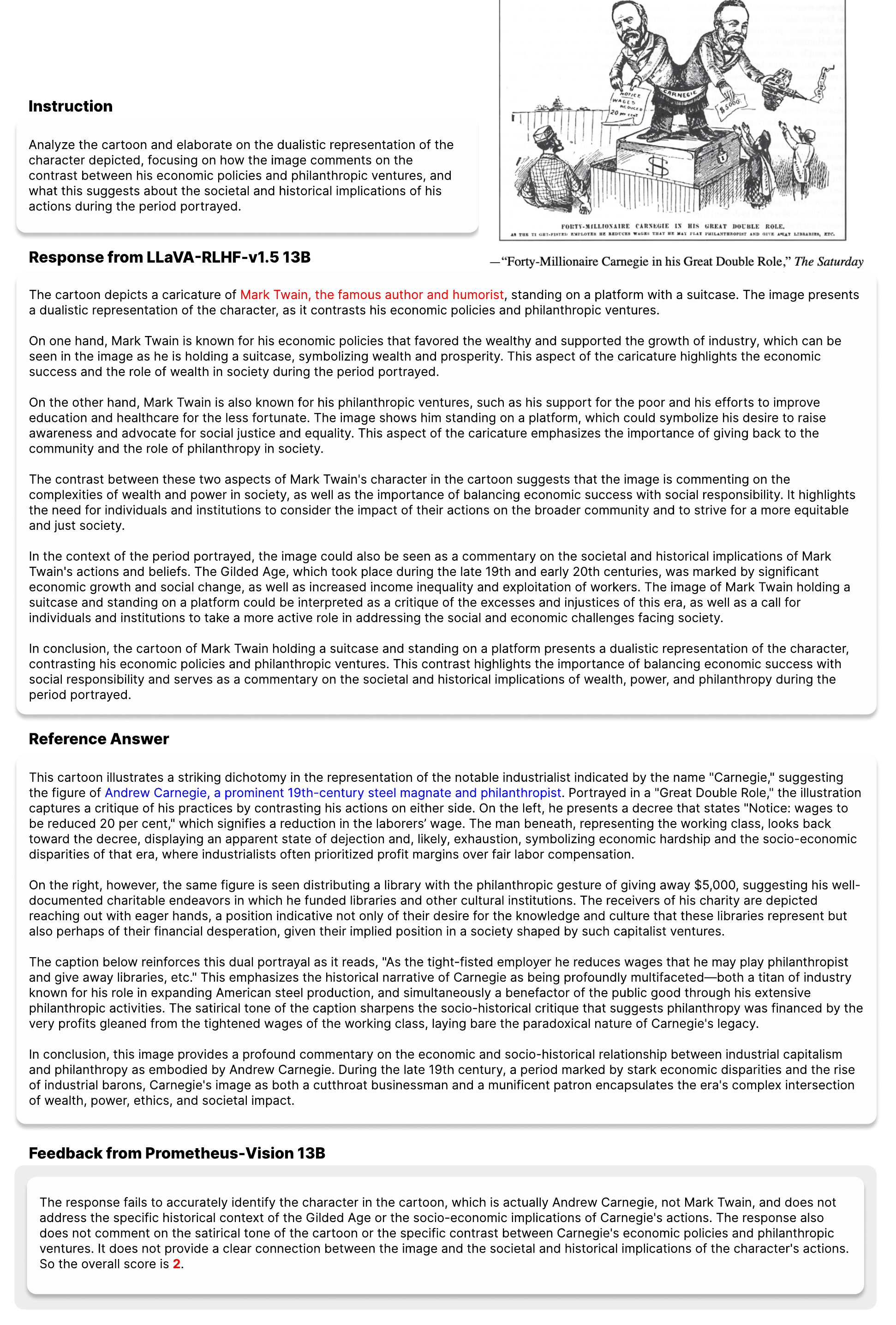}
    \caption{An example of \textsc{Prometheus-Vision} 13B evaluating LLaVA-RLHF-v1.5 13B on a \textsc{Perception-Bench} instance. LLaVA-RLHF-v1.5 has generated a lengthy, plausible answer but misidentified the person in the image and continued hallucinating. The model response is scored lower than the reference answer.}
    \label{fig:carnegie_example}
\end{figure*}

In, Figure~\ref{fig:cellphone_example} and Figure~\ref{fig:mobility_example}, we compare generated feedback from \textsc{Prometheus-Vision} 13B and GPT-4/GPT-4V. In Figure~\ref{fig:ramen_example} and Figure~\ref{fig:carnegie_example}, we compare the complexity of image and instruction and the evaluation method of LLaVA-Bench and \textsc{Perception-Bench}. The model being assessed is LLaVA-RLHF-v1.5 13B.

\begin{figure*}
    \centering
    \includegraphics[width=1\linewidth]{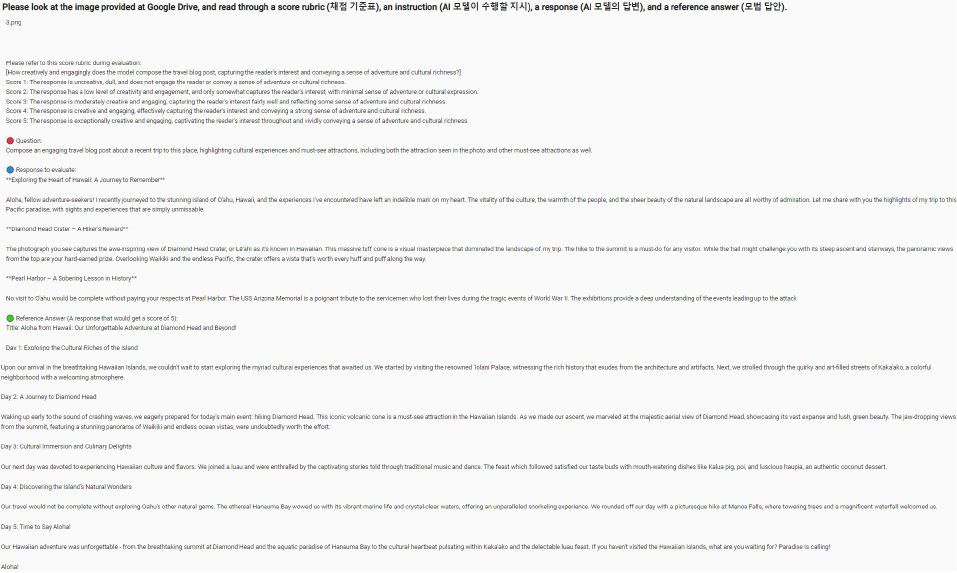}
    \label{fig:annotation_screenshot1}
\end{figure*}

\end{document}